\documentclass[a4paper,fleqn]{cas-dc}
\usepackage{xcolor} 
\usepackage[numbers]{natbib}
\usepackage{float}  
\usepackage{subcaption}
\usepackage{threeparttable}
\usepackage{graphicx} 
\usepackage{amsmath}  
\usepackage{tabularx} 
\usepackage{hyperref}
\usepackage{adjustbox} 

\makeatletter
\@ifundefined{c@dbltopnumber}{}{\setcounter{dbltopnumber}{5}}
\makeatother

\begin{document}
	\def\floatpagepagefraction{1}
	\def\textpagefraction{.01}
	\shorttitle{Ying~Liu et~al. Neurocomputing}
	\shortauthors{Liu et~al.}
	
	\title [mode = title]{Relational graph-driven differential denoising and diffusion attention fusion for multimodal conversation emotion recognition}

	\tnotetext[1]{Our code is available at \href{https://github.com/liuying2023912/ReDiFu}{https://github.com/liuying2023912/ReDiFu}.}
	
	\author[1]{Ying Liu}
	
	\author[1]{Yuntao Shou}
	
	\author[1]{Wei Ai}
	\cormark[1]

	\author[1]{Tao Meng}

	\author[2]{Keqin Li}
	
	\cortext[1]{Corresponding author}
	
	\address[1]{{College of Computer and Mathematics, Central South University of Forestry and Technology, 410004, Hunan, Changsha China}}
	
	\address[2]{{Department of Computer Science, State University of New York, New Paltz, New York, 12561, USA}}

	\begin{abstract}
		Multimodal Conversation Emotion Recognition (MCER) aims to classify the emotional states of utterances by leveraging textual, acoustic, and visual features. However, in real-world scenarios, audio and video signals are often subject to environmental noise and limited acquisition conditions, resulting in extracted features containing excessive noise. Furthermore, there is an imbalance in data quality and information carrying capacity between different modalities. These two issues together lead to information distortion and weight bias during the fusion phase, impairing overall recognition performance. Most existing methods neglect the impact of noisy modalities and rely on implicit weighting to model modality importance, thereby failing to explicitly account for the predominant contribution of the textual modality in emotion understanding.
		To address these issues, we propose a relation-aware denoising and diffusion attention fusion model for MCER. Specifically, we first design a differential Transformer that explicitly computes the differences between two attention maps, thereby enhancing temporally consistent information while suppressing time-irrelevant noise, which leads to effective denoising in both audio and video modalities. Second, we construct modality-specific and cross-modality relation subgraphs to capture speaker-dependent emotional dependencies, enabling fine-grained modeling of intra- and inter-modal relationships. Finally, we introduce a text-guided cross-modal diffusion mechanism that leverages self-attention to model intra-modal dependencies and adaptively diffuses audiovisual information into the textual stream, ensuring more robust and semantically aligned multimodal fusion. Experiments on multiple real-world datasets demonstrate that our model achieves superior performance compared to existing state-of-the-art methods, significantly improving robustness and accuracy in MCER.
	\end{abstract}


	\maketitle
	
	\section{Introduction}
	With the rapid development of artificial intelligence and human-computer interaction technologies, Multimodal Conversation Emotion Recognition (MCER) has emerged as a key research focus in the field of Artificial Intelligence. MCER enables computational systems to gain a deeper understanding of interactive contexts and perceive emotional states in conversations \cite{yan_jiu_fang_xiang, shou2026comprehensive, shou2022conversational}. Due to its broad application prospects in areas such as intelligent customer service \cite{zhi_neng_ke_fu, shou2024adversarial, shou2024low}, virtual assistants \cite{xu_ni_zhu_shou, meng2024deep, shou2025masked}, online education \cite{zai_xian_jiao_yu, meng2024multi, shou2026graph}, and mental health monitoring \cite{xin_li, shou2024efficient, shou2025spegcl}, MCER has received increasing research attention.

\begin{figure}[!t]
		\centering
		\includegraphics[width=\linewidth]{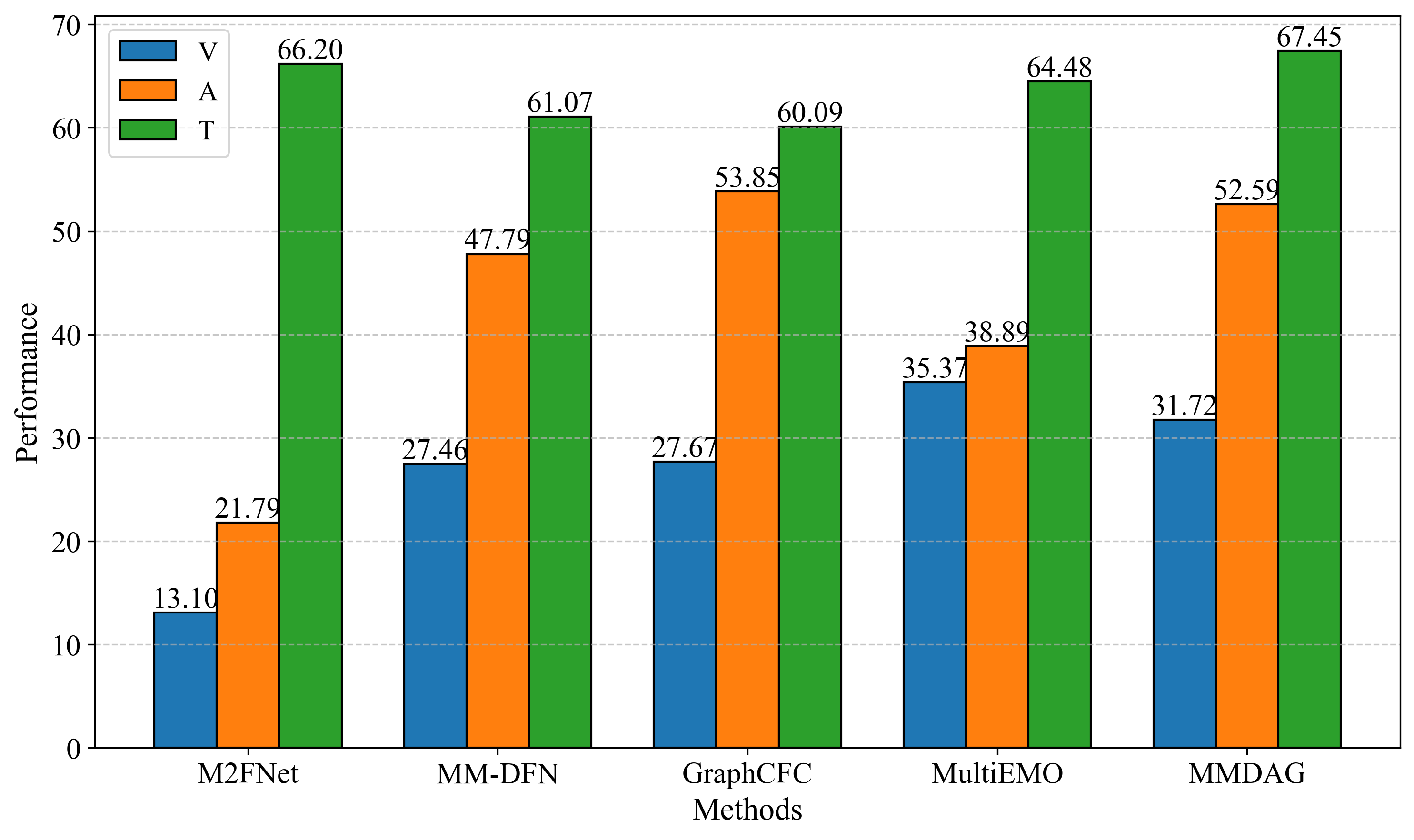}
		\caption{{Performance comparison of various models on the IEMOCAP dataset using unimodal features. T, V, and A denote the textual, visual, and acoustic modalities, respectively.}}
		\label{fig:引言图}
	\end{figure}

	The key to the MCER task lies in modeling semantic dependencies across multiple modalities to achieve complementary feature fusion, thereby enhancing the accuracy of emotion recognition. In recent years, researchers have proposed various fusion strategies to address this challenge. Early concatenation-based fusion methods combined feature vectors from different modalities along the feature dimension to form a unified representation. This representation was then fed into sequential models such as RNNs or LSTMs to learn contextual dependencies for emotion recognition. For instance, Subbaiah et al. \cite{EMRA-Net} proposed the EMRA-Net model, which enhances the concatenated features using a multi-scale residual attention network, followed by temporal modeling using LSTM. However, this method is limited by its simple stacking of modality features, which fails to effectively capture emotional interactions between modalities. Subsequently, Transformer-based fusion methods have become a research hotspot. For example, the JMT model proposed by Waligora et al. \cite{JMT} employs a cross-modal self-attention mechanism to model the correlation weights between modalities. The weighted fused features are then passed through a feedforward neural network for processing. However, the self-attention mechanism may introduce redundant information when capturing global dependencies, which could hinder the recognition of fine-grained emotions. In addition, the semantic gap between modalities may lead to biased attention weights, thereby suppressing information from secondary modalities. GCNs have gradually been adopted in MCER due to their strengths in modeling relationships in non-Euclidean space. The DER-GCN model proposed by Ai et al. \cite{Ai} constructs a weighted multi-relational graph to simultaneously capture interactions between speakers and events. Compared to Transformer-based models, GCNs perform better in handling unstructured multimodal interactions, particularly in scenarios with uneven modality distributions. Although existing methods have achieved remarkable progress in MCER, they still face the following challenges:
	
	\textbf{(i) Neglect of noise interference in audio-visual modalities.}
	In real-world scenarios, noise interference in audio and visual modalities is often far more severe than in the textual modality. For instance, audio signals are frequently disrupted by environmental noise, resulting in blurred prosody, while visual inputs are prone to illumination changes and motion artifacts, which degrade the clarity of facial expressions. Direct fusion of these modalities without noise mitigation introduces interference that makes it difficult for models to accurately capture true emotional variations. However, existing models generally lack explicit mechanisms to handle noise in audio and visual modalities \cite{noise1,noise2,noise3}. For example, Jin et al. \cite{JIN} proposed a Multimodal Graph Transformer that constructs graphs directly using raw audio spectrograms and visual frame data without denoising, which results in node features being derived from noisy inputs. The presence of such noise can lead to spurious connections in the computed adjacency matrix, distorting inter-modal dependencies and consequently degrading recognition performance. Since the impact of noise on audio-visual features is inevitable, and current models lack effective noise-handling mechanisms, significant performance disparities often emerge among different modalities in emotion recognition. Moreover, numerous empirical studies confirm this issue \cite{noisez1,noisez2}. For example, as shown in Figure \ref{fig:引言图}, in the M2FNet model proposed by Chudasama et al. \cite{noise_example}, the accuracy of text-based unimodal emotion recognition reaches 66.20\%, while the accuracy of the visual and audio modalities drops to 13.10\% and 21.79\%, respectively. Therefore, effectively addressing the noise in the audio and visual modalities remains a key challenge in improving the overall emotion recognition performance of multimodal systems.

	\textbf{(ii) Neglect of the dominant role of the textual modality.}
	Most existing multimodal fusion methods model the importance of each modality implicitly through attention weights, often overlooking the dominant role of the textual modality in emotion analysis. For example, Cai et al. \cite{zhudao1} proposed a multimodal sentiment analysis approach that adjusts the contributions of each modality via parameter sharing, but this approach still essentially relies on implicit modeling. However, studies have shown that modality imbalance is an inherent phenomenon in multimodal learning \cite{zhudao2,zhudao3,zhudao4}. As the core carrier of emotional semantics, the textual modality can serve as a guiding modality in the fusion process. Explicitly leveraging text to guide the fusion of other modalities can help mitigate noise interference and enhance overall performance. For instance, in multimodal image fusion, Zhang et al. \cite{zhudao5} proposed a text-modulated diffusion framework in which text guides image fusion, significantly improving image quality and reducing noise. In the field of multimodal fake review detection, Du et al. \cite{zhudao6} introduced a co-attention-based model that emphasizes the dominance of text, thereby enhancing the detection performance. These studies demonstrate that explicitly modeling the dominant role of the textual modality is crucial for optimizing multimodal fusion. Therefore, how to incorporate explicit textual guidance into the fusion process remains another major challenge in MCER.
	
	Based on the above analysis, this paper proposes a novel relation graph-based differential denoising and diffusion attention fusion model for Multimodal Emotion Recognition in Conversations. Our model effectively performs denoising on audio and visual modalities while allowing the text modality to explicitly guide cross-modal fusion, resulting in improved emotion recognition performance. Specifically, we design a differential Transformer module that explicitly computes the differences between two attention maps, effectively enhancing temporally consistent information and thereby suppressing time-irrelevant noise, enabling efficient denoising of the audio and visual modalities. For the text modality, we represent conversations as two subgraphs capturing inter-speaker and intra-speaker emotional dependencies, and assign learnable embeddings to the relation edges. Through incremental interactive learning, the model first captures inter-speaker emotional interactions, followed by intra-speaker emotional dynamics, enabling precise modeling of individual emotional fluctuations. During the fusion stage, we introduce a text-dominant diffusion attention fusion mechanism. First, intra-modal dependencies are captured via self-attention; then, a cross-modal diffusion attention mechanism based on modality similarity allows the text to unidirectionally absorb information from the visual and audio modalities, thereby alleviating the modality gap. The fused representation is then refined via a feedforward network and residual connections, ultimately used for emotion classification. Experimental results demonstrate that our model consistently outperforms existing methods. The key contributions of this study are encapsulated in the following points:
	
	\begin{itemize}
		\item \textcolor{black}{We propose a differential denoising mechanism that enhances dynamic modeling while effectively suppressing noise, significantly improving recognition accuracy.}
		
		\item To address modality imbalance in multimodal fusion, we design a text-dominant diffusion attention fusion mechanism. By explicitly modeling the dominance of the textual modality, our method alleviates the imbalance problem and achieves notable performance gains.
		
		\item Extensive experiments on the IEMOCAP and MELD datasets show that our approach consistently outperforms state-of-the-art baselines.
	\end{itemize}

	
	\section{Related Works}
	MCER is a critical task in affective dialogue systems, aiming to identify emotional states from interactions involving multiple speakers and modalities. This paper reviews and analyzes relevant research from three core perspectives: context-aware modeling, dynamic modeling of emotional dependencies, and Transformer-based multimodal fusion.

	\subsection{Context-Aware Modeling}
	Context-aware modeling within modalities plays a foundational role in MCER.  DialogueRNN \cite{DialogueRNN}, as a pioneering work, first proposed a dedicated recurrent neural architecture to explicitly track speaker-specific emotional states. In recent years, various methods have been proposed to capture dynamic changes within modal sequences. For instance, Self-MM \cite{self-MM} enhances intra-modal context representations through self-supervised learning but lacks explicit modeling of fine-grained frame-level emotional changes. CH-SIMS \cite{CH-SIMS}introduces modality alignment to improve cross-modal consistency, yet its audio and visual modalities rely on static feature extraction, limiting its ability to model continuous temporal variations. CoMPM \cite{CoMPM} designs a collaborative memory module to integrate contextual information, but exhibits limited capability in modeling local emotional fluctuations. EmoCaps \cite{EmoCaps} employs capsule networks for multimodal representation but does not explicitly address intra-modal dynamic variations.

	Recently, MutiEMO \cite{MutiEMO} mainly performs weighted aggregation of historical information via standard self-attention, with a stronger emphasis on context-aggregative representation enhancement. However, it lacks a dedicated explicit constraint on abrupt temporal fluctuations in the attention distribution matrix. In contrast, we introduce a differential attention mechanism that explicitly compares the current attention distribution with a temporally shifted reference attention distribution, and further suppresses temporally inconsistent pseudo-relations through gated filtering. This design allows the model to produce a relational change response while performing context aggregation, thereby suppressing perceptual noise and more accurately capturing emotion-related transient dynamics.

	\subsection{Modeling Dynamic Emotional Dependencies}
	Graph-structured temporal modeling has been widely used to characterize complex dynamic dependencies and has been validated across various sequential prediction tasks \cite{uygun2025financial,celik2025analyzing, ai2026paradigm, shou2025contrastive, shou2025revisiting}. In MCER, speaker interactions likewise exhibit pronounced relational structural properties; therefore, graph-based modeling has become a common approach for capturing inter-speaker emotional dependencies. DialogueGCN \cite{DialogueGCN} introduces graph convolutional networks and defines multiple relation types (e.g., self, inter-speaker, and temporal relations) to model conversational structure. However, such single-conversation graph methods typically perform message passing in a unified graph, without explicitly separating inter-speaker interactions and intra-speaker emotional inertia into independent subgraphs. As a result, it is difficult to clearly distinguish the contributions of these two types of dependencies at the structural level. COGMEN \cite{COGMEN} enhances emotional reasoning through cognitive graphs but does not distinguish between self-dependency and inter-speaker influence, resulting in coarse structural granularity. AGF-IB \cite{AGF-IB} employs adversarial alignment and graph contrastive learning to fuse multimodal features, yet its graph focuses solely on cross-modal interactions without explicitly modeling the coupling between speaker identity and contextual information. MKE-IGN \cite{MKE-IGN} introduces a multi-knowledge enhanced interaction graph network by encoding the relations between utterances and commonsense knowledge via edge representations, but it fails to differentiate speaker-to-speaker interactions from intra-speaker emotional inertia, leading to ambiguous emotion propagation paths. Although SAM+GCN \cite{SAM+GCN} demonstrates the potential of GCNs for static visual recognition, it primarily emphasizes spatial correlations rather than the dynamic temporal evolution in conversations. DGODE \cite{DGODE} proposes a dynamic graph neural ODE network that leverages continuous-time modeling and mixed-hop mechanisms to characterize emotion dynamics as context evolves, effectively alleviating the over-smoothing problem in GCNs. 
	
	In contrast, we construct independent inter-speaker (InterGAT) and intra-speaker (IntraGAT) interaction subgraphs, decoupling the two types of dependencies into separe and controllable propagation paths. We further employ graph attention to adaptively learn the importance of edges, enabling finer-grained modeling of dynamic emotional dependencies. Therefore, while preserving the interpretability of the graph structure, our method is better suited to characterize complex speaker interaction patterns and emotion evolution processes.
	
	\subsection{Multimodal Fusion Transformer}
	The key challenge in multimodal fusion lies in effectively harmonizing heterogeneous information across different modalities \cite{shou2025dynamic, shou2025gsdnet, shou2023graphunet, shou2025graph, shou2025multimodal, shou2025graph, shou2025cilf}. Recently, MulT \cite{MulT} proposed a cross-modal Transformer that achieves temporal alignment via cross-attention but lacks sufficient modeling of modality structural dependencies. MISA \cite{MISA} distinguishes between shared and private modality representations to reduce redundancy. However, its fusion mainly relies on feature concatenation without structural guidance. MAG-BERT \cite{MAG-BERT} introduces modality augmentation based on textual features to enhance non-verbal modalities, yet it does not explicitly model cross-modal interaction structures. Self-MM optimizes fusion consistency through contrastive learning, but the fused representations remain susceptible to modality conflicts and instability. MM-DFN \cite{MM-DFN} proposes a dynamic fusion network, yet it essentially remains an implicit fusion paradigm. Moreover, although DialogueTRM \cite{DialogueTRM} exhibits strong temporal modeling capability, it does not incorporate a reliable modality (e.g., text) as a semantic anchor, and thus its cross-modal alignment can be insufficiently stable under noisy conditions. 
	
	In contrast, we first suppress temporally inconsistent noise at the interaction level via a differential mechanism; we then introduce text-dominant diffusion fusion, where the semantically reliable textual modality serves as an anchor to explicitly guide semantic alignment. Different from conventional weighted fusion schemes, our design structurally integrates noise blocking with semantic synergy, thereby substantially improving the robustness of the fused representations.
	
	Despite the progress made by existing methods in modeling the dynamic nature and structural consistency of multimodal emotions, there remains a lack of effective denoising strategies for the audio and visual modalities. Moreover, current fusion approaches typically rely on weight-based implicit modeling, which limits their ability to capture collaborative relationships among modalities. We propose a differential denoising mechanism and incorporate a text-dominant cross-modal diffusion strategy at the fusion stage to enhance the robustness and semantic consistency of the fused representation.

	\section{Methodology}
\begin{figure*}[!t]
		\centering
		\includegraphics[width=0.99\linewidth]{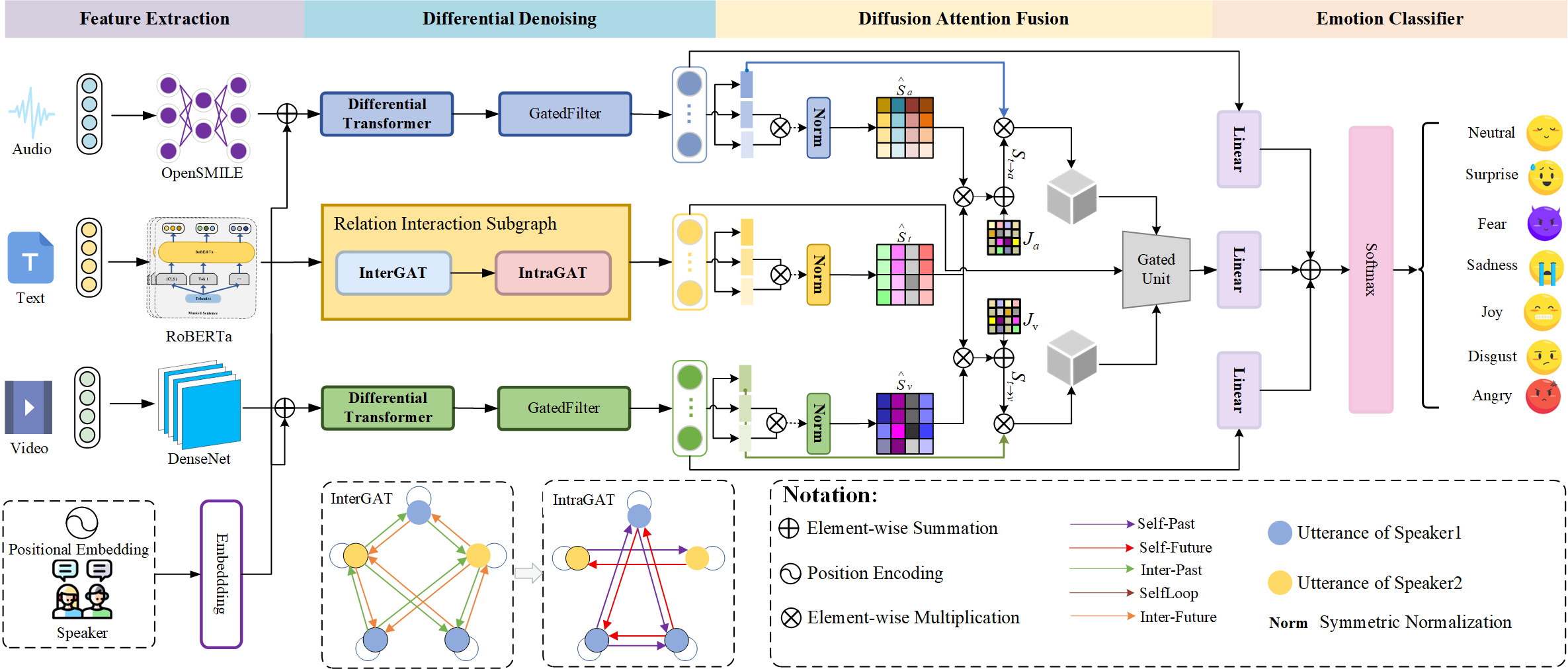}
		\caption{The overall framework is as follows: first, extract utterance-level unimodal features from audio, video, and text; then, apply a differential Transformer to audio and video features for dynamic extraction and denoising, while modeling emotional dependencies of textual features through a relational subgraph module comprising InterGAT and IntraGAT; finally, perform emotion classification by fusing multimodal features with text as the core using cross-modal diffusion attention.}
		\label{fig:整体架构图}
	\end{figure*}

	The core objective of MCER is to allocate suitable emotion labels to every individual utterance contained in a dialogue. To address the challenges of noise interference in the audio and visual modalities, as well as information fusion bias caused by modality imbalance, this paper proposes a novel framework: Relational Graph-Driven Differential Denoising and Diffusion Attention Fusion. Specifically, a differential Transformer is employed to dynamically model the audio and visual modalities while effectively suppressing noise. In parallel, intra-speaker and inter-speaker relational graphs are constructed to capture emotional dependencies within the textual modality. Furthermore, a text-guided cross-modal diffusion attention fusion mechanism is introduced to enable effective integration of multimodal emotional information. Figure \ref{fig:整体架构图} shows the proposed framework.
	
	\subsection{Task Definition}
	In the MCER task, a conversation consists of a sequence of utterances \( U=\{u_1, u_2, \cdots, u_N\} \) and a sequence of speakers \( S = \{s_1, s_2, \cdots, s_M\} \), where $N$ is the utterance count and $M$ is the speaker count. Each utterance \( u_i \) contains three modalities of information: text (\( t \)), audio (\( a \)), and visual (\( v \)). The goal of this paper is to predict the emotion category of the speaker at the current time step \( q \) based on the speaker's text, audio, and visual features. The emotion prediction task is defined as follows:
	\begin{equation}
		y_i = \text{prediction}(\{u_1, u_2, \cdots, u_i\}), \quad i \in [q - k, q - 1],
	\end{equation}
	where $y_i$ refers to the emotional annotation of utterance $i$, and $k$ determines the length of the preceding context considered.

	\subsection{Feature Extraction }
	\textbf{Text Feature Extraction:} Following previous studies \cite{Feature_Extraction1, Feature_Extraction2}, we employ the RoBERTa model to obtain contextual representations of the input text. Specifically, the input sequence is tokenized and augmented with special tokens [CLS] and [SEP], then fed into the RoBERTa model. The feature representation for the text modality is derived from the output associated with the [CLS] token in the final layer.
	
	\textbf{Visual and Audio Feature Extraction:} In line with prior work \cite{Feature_Extraction3, Feature_Extraction1}, we utilize a pre-trained DenseNet model and the openSMILE toolkit to extract visual and audio features, respectively.
	
	Due to the inconsistency in the original feature dimensions across the text, audio, and visual modalities, we introduce independent linear projection layers for each modality to map them into a common feature space for subsequent alignment processing:
	\begin{equation}
		Z^m = W^m \cdot H^m + b^m, \quad m \in \{t, a, v\},
	\end{equation}
	where \( H^m \) represents the original modality features, and \( W^m \), \( b^m \) are trainable weights and biases corresponding to modality \( m \in \{t, a, v\} \).

	\subsection{Speaker and Position Embeddings}
	To distinguish speakers in a conversation and model their individual features, we encode identity information into learnable vector representations through speaker embeddings. Specifically, for the \(i\)-th utterance \(u_i\) spoken by speaker \(s_j\), its identity is first represented as a one-hot vector \(O(s_j)\), and then transformed into an embedding vector \(spk_i = E_{spk} \cdot O(s_j) \in \mathbb{R}^d\) via a learnable speaker embedding matrix \(E_{spk} \in \mathbb{R}^{M \times d}\), where $M$ represents the total count of speakers and $d$ specifies the dimension of the embedding space.
	The speaker embedding vectors of all utterances form a sequence:
	\begin{equation}
		SE = [spk_1, spk_2, \cdots, spk_N] \in \mathbb{R}^{N \times d}.
	\end{equation}
	
	To model the temporal structure of utterances, we adopt sinusoidal positional encoding to provide temporal position information. For an utterance at position \(pos\) in the sequence, its positional encoding is calculated as follows:
	\begin{align}
		PE(pos, 2i) &= \sin\left(\frac{pos}{10000^{2i/d}}\right), \\
		PE(pos, 2i+1) &= \cos\left(\frac{pos}{10000^{2i/d}}\right),
	\end{align}
	where \(i\) is the index of the feature dimension, \(d\) is the embedding dimension, and \(PE \in \mathbb{R}^{N \times d}\) is the positional encoding matrix for all utterances.
	
	Finally, we obtain the fused representation through element-wise summation:
	\begin{equation}
		\hat{Z}_{in}^m = Z^m + SE + PE \in \mathbb{R}^{N \times d}, \quad m \in \{a, v\}.
	\end{equation}

	\subsection{Differential Denoising}
	During audio and visual acquisition and encoding, background noise, static scene components, and speaker-/scene-related biases are often unavoidable, which introduces emotion-irrelevant redundancy into the extracted features. From the perspective of attention-based relations, we posit that emotion-discriminative cues are more likely reflected in the variations of relational patterns over a local temporal span, whereas stationary or slowly varying disturbances tend to keep the attention distributions at adjacent time steps similar. Based on this assumption, we perform the differencing operation in the relation attention domain rather than directly applying temporal differencing to feature values. This design attenuates stationary relational redundancy while preserving the original semantic information, and yields stronger responses to more discriminative dynamic changes. Meanwhile, the differential mechanism is also sensitive to abrupt perturbations; therefore, we further introduce gated filtering to reduce the influence of random jumps caused by non-stationary noise, thereby improving overall robustness.
	
	\subsubsection{Differential Transformer}
	Given the input sequence representation of modality $m \in \{a, v\}$, denoted as $\hat{Z}^m_{in} \in \mathbb{R}^{N \times d}$, we first apply linear projections to obtain the query, key, and value:
	\begin{equation}
		Q = \hat{Z}^m_{in} W_Q, \quad K = \hat{Z}^m_{in} W_K, \quad V = \hat{Z}^m_{in} W_V,
		\label{eq:7}
	\end{equation}
	where $W_Q, W_K, W_V \in \mathbb{R}^{d \times d}$ are learnable parameter matrices.
	
	To construct a temporal reference view within the same sequence, we build a reference \text{key} via a first-order temporal shift. Specifically, we set $K_t^{ref} = K_{t-1} \ (t > 1)$ and adopt a replication strategy at the boundary $t = 1$, i.e., $K_1^{ref} = K_1$. We then compute attention distributions using the current key and the reference key, respectively:
	\begin{equation}
		\alpha = \text{softmax}\left( \frac{Q K^\top}{\sqrt{d}} \right), \quad \alpha^{ref} = \text{softmax}\left( \frac{Q (K^{ref})^\top}{\sqrt{d}} \right),
		\label{eq:8}
	\end{equation}
	where $\text{softmax}(\cdot)$ normalizes along the key dimension, such that $\alpha$ and $\alpha^{ref}$ provide, for each query, an association distribution over all time steps; $\sqrt{d}$ is a scaling factor used to stabilize dot-product magnitudes and gradients.
	
	Based on the two distributions, we take their difference to characterize relational changes, and introduce a learnable scalar $\lambda$ to adjust the contrast strength:
	\begin{equation}
		\text{DiffAttn}(\hat{Z}^m_{in}) = (\alpha - \lambda \alpha^{ref})V,
		\label{eq:9}
	\end{equation}
	where $\lambda$ is a learnable scalar that adaptively controls the cancellation strength of the reference view; the term $(\alpha-\lambda\,\alpha^{ref})$ is no longer a probability distribution, but serves as a response weight to relational changes.

	To enhance representation capability, we adopt a multi-head mechanism. The input is projected into $h$ subspaces and differential attention is computed independently in each, all sharing the same $\lambda$. Outputs from each head are normalized via GroupNorm, concatenated, and fused via a linear transformation:
	\begin{align}
		\mathrm{head}_i &= \mathrm{DiffAttn}(\hat{Z}_{in}^m; W_{Q_i}, W_{K_i}, W_{V_i}, \lambda)
		\label{eq:10},   \\
		\check{Z}_{attn}^m &= \mathrm{Concat}(\mathrm{head}_1, ..., \mathrm{head}_h) W_O,
	\end{align}
	where $W_{Q_i}, W_{K_i}, W_{V_i}$ are the projection matrices for the $i$-th head, and $W_O$ is the output fusion matrix.

	To avoid damaging stationary yet useful semantic components during filtering, we retain a residual path to stabilize the representation learning process:
	\begin{equation}
		\tilde{Z}_{res}^m = \hat{Z}^m_{in} + \check{Z}_{attn}^m.
	\end{equation}
	
	Compared with the commonly used temporal differencing in the feature-value domain, e.g., $Z_t - Z_{t-1}$, this module directly models changes in relational patterns at the level of attention distributions and can thus more directly suppress stationary relational redundancy. Moreover, it does not rely on an additional contrastive-learning objective; instead, it constructs an explicit contrast through a temporal reference within the same sequence. Compared with generic delta-attention incremental update formulations, the relational change signal in our design is realized by directly differencing two attention distributions, which admits a clear interpretation as a relational change response.
	
	\subsubsection{Gated Filtering}
	The differential mechanism responds to relational changes; such responses may arise from emotion-relevant dynamics or be triggered by instantaneous perturbations of non-stationary noise. To mitigate the instability introduced by the latter, we introduce element-wise gating on the residual output:
	\begin{equation}
		F = \sigma({Z}_{res}^m \tilde W_g  + b_g)
		\label{eq:12},
	\end{equation}
	where $W_g$ and $b_g$ are learnable parameters, and $\sigma$ denotes the Sigmoid function ensuring the gating weights lie in the range $[0, 1]$.
	
	Finally, the gated feature is obtained by element-wise multiplication between the residual output and the gate:
	\begin{equation}
		Z'_m = \tilde{Z}_{res}^m \odot F,
		\label{eq:13}
	\end{equation}
	where $\odot$ denotes element-wise multiplication.

	\subsection{Relation Interaction Subgraph}
	\subsubsection{Graph Construction}
	
	To capture the dynamic emotional dependencies both between speakers and within individual speakers in multi-party dialogues, this paper constructs two subgraphs: the inter-speaker emotional interaction subgraph \( G_{inter} \) and the intra-speaker emotional interaction subgraph \( G_{intra} \). The dialogue consists of \( N \) ordered utterances, each corresponding to a node \( v_i \) with temporal index \( i \) and initial features represented by the textual feature vector \( z_i^t \in \mathbb{R}^d \).
	
	Edges in these two subgraphs are constructed based on a time-window mechanism, connecting only nodes whose temporal index difference does not exceed the window size \( k \), in order to reduce noise interference from long-range contexts. Under this rule, the edge topology of the two subgraphs is predefined and static, and the presence of an edge is determined solely by the time indices and speaker identities. In the inter-speaker subgraph, edges are established only when the temporal index difference between two nodes is at most \( k \) and the utterances come from different speakers. In the intra-speaker subgraph, edges are created only when the temporal index difference is at most \( k \) and the utterances originate from the same speaker. Formally, the edge definitions are:
	\begin{align}
		e_{ij}^{inter} &=
		\begin{cases}
			1, & |i - j| \leq k, \text{ and } \text{speaker}(i) \neq \text{speaker}(j), \\[4pt]
			0, & \text{otherwise}
		\end{cases},\\
		e_{ij}^{intra} &=
		\begin{cases}
			1, & |i - j| \leq k, \text{ and } \text{speaker}(i) = \text{speaker}(j), \\[4pt]
			0, & \text{otherwise}
		\end{cases}.
	\end{align}
	
	Edges in both subgraphs include three types: self-loops (when \( i = j \)) indicating a node's dependence on its own emotional state; forward edges (when \( i < j \)) representing the influence of past utterances on future ones; and backward edges (when \( i > j \)) reflecting the reverse emotional feedback from future utterances to past ones. To systematically model the temporal emotional dynamics, we introduce a relation type label \( r_{ij} \in \{1, 2, 3\} \) corresponding to self-loop, forward, and backward edges respectively.
	
	To facilitate learning of edge weights, a relation embedding matrix \( R_e \in \mathbb{R}^{N_r \times d_r} \) is designed, where \( N_r \) is the number of relation types and \( d_r \) is the dimension of the relation embedding. Edge features are obtained by multiplying the one-hot encoded vector of the relation type \( O(r_{ij}) \) with the relation embedding matrix:
	\begin{equation}
		e_{ij} = R_e \cdot O(r_{ij}) \in \mathbb{R}^{d_r}.
	\end{equation}

	\subsubsection{Graph Interaction}
	
	We use a Graph Attention Network (GAT) for message passing on each subgraph, where the edge attention weights $\alpha_{ij}$ are learned during training to adaptively aggregate neighborhood information. We further incorporate relation-type embeddings $e_{ij}$ into the attention computation, allowing the model to distinguish different conversational relations (e.g., self-loop, forward, and backward edges). The edge attention weight is computed as follows:
	\begin{equation}
		\alpha_{ij} = \text{softmax} \big( \text{LeakyReLU} \big( a^\top [W z_i \| W z_j \| e_{ij}] \big) \big),
	\end{equation}
	where \( z_i \) and \( z_j \) denote the feature representations of nodes \( v_i \) and \( v_j \) at time step \( t \), \( W \) is a learnable weight matrix, \( e_{ij} \) is the edge feature, \( \| \) denotes vector concatenation, and \( a \) is a learnable attention vector.
	
	Based on the computed attention weights, the hidden feature of node \( v_i \) is updated as:
	\begin{equation}
		z_{i, \text{out}}^t = \sigma \left( \sum_{j \in \mathcal{N}_i} \alpha_{ij} W z_j^t \right),
	\end{equation}
	where \( \mathcal{N}_i \) represents the neighbor set of node \( v_i \), \( \sigma \) is a nonlinear activation function, and $z_j^t$ represents the features of neighbor node $v_j$ at time $t$.
	
	Finally, the textual modality feature representation is formed by concatenating the updated feature vectors of all nodes:
	\begin{equation}
		Z_{\text{out}}^t = [z_{1, \text{out}}^t, z_{2, \text{out}}^t, \dots, z_{N, \text{out}}^t] \in \mathbb{R}^{N \times d}.
	\end{equation}

	\subsection{Diffusion Attention Fusion}
	To address the impact of modality imbalance on the performance of multimodal fusion, this paper proposes a text-centric cross-modal diffusion attention fusion mechanism. This mechanism leverages the dominant role of the text modality in sentiment analysis by explicitly constructing an interaction framework between the text and other modalities. Information transfer is achieved through cross-modal correlations, and a gating strategy is incorporated to dynamically fuse multi-source features, thereby enhancing the effectiveness of multimodal fusion.

	First, the features of the audio, text, and visual modalities are each linearly transformed and projected into a unified representation space, obtaining the query (\(Q\)), key (\(K\)), and value (\(V\)) matrices. Then, the self-attention matrix is computed to capture intra-modal feature correlations:
	\begin{equation}
		S_m = \mathrm{Softmax}\left(\frac{Q_m K_m^\top}{d}\right), \quad m \in \{a, t, v\},
	\end{equation}
	where \(d\) is a scaling factor that prevents the dot product values from becoming too large, thus avoiding gradient vanishing. 
	
	Since the attention matrices of different modalities have distinct value distributions, directly computing cross-modal correlations may cause weight bias or information loss in certain modalities. To ensure the stability and symmetry of cross-modal correlations, the self-attention matrix is normalized by the degree matrix:
	\begin{equation}
		\hat{S}_m = D_m^{-\frac{1}{2}} S_m D_m^{-\frac{1}{2}},
	\end{equation}
	where \(D_m\) denotes the degree matrix of \(S_m\).
	
	To enable the text modality to learn useful information from the audio and visual modalities, 
	the cross-modal attention matrix is constructed by combining the normalized correlations 
	with the original self-attention matrices:
	\begin{equation}
		S_{t \to m} = \gamma \cdot \left( \hat{S}_t \hat{S}_m^\top \right) + (1 - \gamma) \cdot J_m, 
		\quad m \in \{a, v\},
		\label{eq:fusion_weight}
	\end{equation}
	where $\gamma \in [0,1]$ is a balancing coefficient that weighs the contribution of the two types 
	of correlation information, and $ J_m = S_t + S_m $ represents the sum of the original self-attention matrices.
	
	Based on the cross-modal attention matrix, the value matrix of the other modality is mapped into the text space to enable information transfer and fusion:
	\begin{equation}
		U_{t \to m} = S_{t \to m} V_m, \quad m \in \{a, v\}.
	\end{equation}
	
	To adaptively adjust the fusion ratio of different modal information, a gating mechanism is designed to dynamically compute the fusion weights based on the transferred features:
	\begin{equation}
		T_t = \sigma \left( W_g \left[ U_{t \to a} \parallel U_{t \to v} \right] \right),
	\end{equation}
	where \(\sigma\) is the Sigmoid activation function with output range \([0,1]\), representing the weight of information received from the audio modality; \(1 - T_t\) corresponds to the weight from the visual modality; \(\parallel\) denotes vector concatenation; and \(W_g\) is a learnable weight matrix.
	
	Finally, the fused text feature is composed of the original text output and the weighted audio and visual features:
	\begin{equation}
		Z_t' = Z_{out}^t + T_t \odot U_{t \to a} + (1 - T_t) \odot U_{t \to v},
	\end{equation}
	where \(\odot\) denotes element-wise multiplication.

	\subsection{Emotion Classifier}
	

	In emotion classification, the classification scores for each modality are first computed through a linear transformation:
	\begin{equation}
		L_m = \mathrm{Linear}_m(Z_m') \in \mathbb{R}^{N \times c}, \quad m \in \{a, t, v\},
	\end{equation}
	where \(c\) denotes the number of emotion categories.
	
	Then, to integrate information from different modalities, the scores from the three modalities are combined by a weighted summation to obtain the final score:
	\begin{equation}\label{27}
		Z_f = \alpha_a L_a + \alpha_t L_t + \alpha_v L_v \in \mathbb{R}^{N \times c},
	\end{equation}
	where \(\alpha_t\), \(\alpha_a\), and \(\alpha_v\) are the weighting coefficients controlling the influence of each modality on the final prediction.
	
	Finally, to obtain the emotion classification probabilities for each sample, the final scores are normalized by a softmax function:
	\begin{equation}
		\hat{Y} = \mathrm{softmax}(Z_f) \in \mathbb{R}^{N \times c},
	\end{equation}
	where \(\hat{Y}\) represents the probability distribution over emotion categories for each sample.


	To improve the consistency of multimodal feature representation and the stability of training, we introduce a multi-constraint joint loss function, which consists of two parts: the emotion fusion loss \(L_f\) and the modality self-supervised loss \(L_u\).
	\begin{equation}
		L_f = -\frac{1}{N} \sum_{i=1}^{N} \sum_{j=1}^{C} y_{i,j} \log(\hat{y}_{i,j}^f),
	\end{equation}
	where $N$ is the total number of samples and $C$ is the total emotion categories, \( y_{i,j} \in \{0, 1\} \) is the one-hot encoded true label, and \( \hat{y}_{i,j}^f \in [0, 1] \) denotes the predicted probability that the \( i \)-th sample belongs to the \( j \)-th category based on the fused multimodal features.
	
	\begin{equation}
		L_u = -\frac{1}{N} \sum_{m \in \{a, v, t\}} \lambda_m \sum_{i=1}^{N} \sum_{j=1}^{C} y_{i,j} \log(\hat{y}_{i,j}^m),
	\end{equation}
	where \( \hat{y}_{i,j}^m \) represents the predicted probability that the \( i \)-th sample belongs to the \( j \)-th category based on modality \( m \) (audio, visual, or text) features, and \( \lambda_m \) is the learning weight factor for modality \( m \), dynamically set to 0.1 times the current classification loss for that modality, in order to prioritize learning the modality with poorer classification performance.
	
	Finally, the total loss \( L_s \) is the sum of the emotion fusion loss and modality self-supervised loss:
	\begin{equation}
		L_s = L_f + L_u.
	\end{equation}

	\section{EXPERIMENTAL SETTINGS}
	
	In this section, we first provide a comprehensive description of the two public datasets used in the experiments and the two key evaluation metrics for model performance. Next, we introduce the baseline models and technical details. Finally, we detail the implementation specifics of the experiments.
	
	\subsection{Datasets}
	
	To evaluate our model, we conduct experiments on the IEMOCAP and MELD datasets, and Table \ref{table:dataset_stats} lists their key statistics.

	\begin{table}[!t]
		\centering
		\caption{The statistics of IEMOCAP and MELD.}
		\label{table:dataset_stats}
		\begin{adjustbox}{width=\linewidth} 
        \setlength{\tabcolsep}{12pt}{
			\begin{tabular}{c|cc|cc|c}
				\hline
				\multirow{2}{*}{Dataset} & \multicolumn{2}{c|}{Dialogues} & \multicolumn{2}{c|}{Utterances} & \multirow{2}{*}{Classes} \\ \cline{2-5}
				& Train\&Valid      & Test      & Train\&Valid       & Test       &                          \\ \hline
				IEMOCAP                  & 120               & 31        & 5810               & 1623       & 6                        \\ 
				MELD                     & 1152              & 280       & 11098              & 2610       & 7                        \\ \hline
			\end{tabular}}
		\end{adjustbox}
	\end{table}

	\textbf{IEMOCAP} \cite{IEMOCAP}  is a multimodal emotional dataset released by the University of Southern California in 2008, focusing on dyadic interaction scenarios. It contains 553 dialogue segments performed by 10 actors (5 male-female pairs), covering both scripted and improvised dialogues. The dataset includes speech, video, and text transcription modalities, annotated with six core emotions: happy, sad, angry, frustrated, excited, and neutral, as well as some emotional intensity. It has high data quality and strong annotation consistency, making it suitable for cross-modal emotion recognition research in natural interactions.
	
	\textbf{MELD} \cite{MELD}, based on the TV show \textit{Friends}, was released in 2018 and focuses on multi-party dialogue scenarios. It includes 1433 dialogue segments and 13708 dialogue units, involving 3-4 person interactions. The dataset provides text, speech, and video modalities (with some versions containing only text and audio) and is annotated with seven emotions: joy, sadness, anger, fear, surprise, disgust, and neutral, along with sentiment polarity. The dataset is large in scale, has complex scenes, and strong context dependencies, making it suitable for multimodal emotion recognition tasks involving long context dependencies.
	
	\subsection{Evaluation Metrics}
	
	We select the weighted accuracy (w-Acc) and weighted F1-score (w-F1) to evaluate model performance. The w-Acc provides an intuitive representation of the model’s overall performance and is easy to compute, making it suitable for comprehensive evaluation when the dataset is balanced. It is calculated as follows:
	\begin{equation}
		w\text{-}Acc = \sum_{i=1}^{C} \frac{N_i}{N} \times Acc_i,
	\end{equation}
	where $C$ is the total number of emotion categories, $N_i$ is the number of samples in the $i$-th class, $N$ is the total number of samples across all classes, and $Acc_i$ denotes the prediction accuracy for the $i$-th class. 
	
	The
	w-F1, on the other hand, balances the class imbalance problem and reflects the model’s accuracy and coverage of each
	emotion, especially being more sensitive to rare emotions. The calculation formula is:
	\begin{equation}
		w\text{-}F1 = \sum_{i=1}^{C} \left( \frac{N_i}{N} \times F1_i \right),
	\end{equation}
	where $F1_i$ is the F1-score for the $i$-th class, defined as:
	\begin{equation}
		F1_i = \frac{2 \times P_i \times R_i}{P_i + R_i},
	\end{equation}
	where $P_i$ and $R_i$ represent the Precision and Recall for the $i$-th class, respectively. By integrating precision and recall while accounting for the weight of each class, w-F1 provides a more objective assessment for imbalanced datasets.
	
	\subsection{Baseline Models}
	
	We conducted a comparative evaluation of multiple baseline models on the IEMOCAP and MELD datasets, including DialogueRNN \cite{DialogueRNN}, MMGCN \cite{MMGCN}, DialogueTRM \cite{DialogueTRM}, MM-DFN \cite{MM-DFN}, DialogueGCN \cite{DialogueGCN}, TS-GCL \cite{wx11}, GraphCFC \cite{GraphCFC}, MultiEMO \cite{MutiEMO}, SDT \cite{wx7} and DEDNet \cite{wx12}.

	\subsection{Implementation Details}
	
	The proposed model is implemented in PyTorch and trained with the Adam optimizer using a cosine-annealing learning-rate schedule. All hyperparameters are selected on the validation set via a restricted grid search with weighted-F1 as the primary criterion, while fixing the encoder depth and batch size to ensure fair comparisons. To address the dimensional discrepancies among textual, acoustic, and visual features on IEMOCAP and MELD, we linearly project each modality into a unified hidden space. The hidden size is set to 512 for IEMOCAP and 256 for MELD; accordingly, the Transformer uses 64 attention heads on IEMOCAP and 16 on MELD to keep the per-head dimension $d_h=d/h$ within a stable range. The context window length is set to 20 for IEMOCAP and 25 for MELD, with initial learning rates of $1\times10^{-4}$ and $5\times10^{-5}$, respectively.

	The architecture integrates a Transformer encoder and a graph attention network. The graph branch uses two attention heads and introduces a 10-dimensional relation embedding to model speaker-dependent relations; the relation embedding dimension is also selected on the validation set. The learnable coefficient $\lambda$ in the differential attention is initialized to 0.5 and updated during training to balance denoising strength and semantic preservation. Multimodal fusion is implemented with MutualFormer, which serves as a mutual-attention-based cross-modal interaction module that enables each modality to learn complementary information from the other modalities and adaptively combines features through a gating mechanism. We set an MLP expansion factor of 4 and a DropPath rate of 0.1 (i.e., stochastic-depth regularization, which randomly drops residual paths during training to mitigate overfitting and improve generalization). To mitigate overfitting, we apply dropout at multiple levels of the network, along with L2 weight decay and layer normalization. The training objective is a class-weighted masked negative log-likelihood loss, where masking is used to exclude padded positions from loss computation, and class weights are used to mitigate training bias caused by class imbalance.
	
	At the feature level, cross-modal complementarity is adaptively modeled by the proposed diffusion and gating modules; at inference time, we further calibrate the modality-specific classification scores via a fixed, validation-tuned weighted fusion. The fusion weights are set to 1.0/1.0/0.4 for IEMOCAP and 3.0/1.0/0.3 for MELD (text/audio/visual).

	\section{EXPERIMENTAL RESULTS}

	In this section, we present and analyze the findings from comparative experiments and ablation studies.
	
	\subsection{Main Results}
	
	As shown in Tables \ref{tab:iemocap_results} and \ref{tab:meld_results}, we compare the performance of our model with baseline methods on the IEMOCAP and MELD datasets. On IEMOCAP, our model achieves a w-Acc of \textbf{75.17\%} and a w-F1 of \textbf{74.87\%}; on MELD, it attains a w-Acc of \textbf{66.52\%} and a w-F1 of \textbf{66.62\%}. All results are averaged over five independent runs using different random seeds. Compared to state-of-the-art baselines, our model demonstrates significant improvements, validating the effectiveness of the proposed differential denoising module and text-dominant cross-modal diffusion attention fusion module. Our approach effectively reduces noise interference and enhances feature purity, while the text-guided fusion strategy strengthens the complementarity and consistency of multimodal information, significantly improving emotion recognition accuracy and robustness.

	\begin{table*}[!t]
		\centering
		\caption{Performance on the IEMOCAP dataset.  * indicates baseline models re-implemented using our extracted features. Bold values indicate the best performance.}
		\label{tab:iemocap_results}
		\resizebox{\linewidth}{!}{
			\begin{tabular}{l|*{12}{c}|cc}
				\toprule
				\multirow{4}{*}{Models} & \multicolumn{12}{c|}{IEMOCAP} & \multirow{4}{*}{w-Acc} & \multirow{4}{*}{w-F1} \\
				\cmidrule(lr){2-13}
				& \multicolumn{2}{c}{happy} & \multicolumn{2}{c}{sad} & \multicolumn{2}{c}{neutral} & \multicolumn{2}{c}{angry} & \multicolumn{2}{c}{excited} & \multicolumn{2}{c|}{frustrated} &  &  \\
				\cmidrule(lr){2-3}\cmidrule(lr){4-5}\cmidrule(lr){6-7}\cmidrule(lr){8-9}\cmidrule(lr){10-11}\cmidrule(lr){12-13}
				& ACC & F1 & ACC & F1 & ACC & F1 & ACC & F1 & ACC & F1 & ACC & F1 &  &  \\
				\midrule
				DialogueRNN & 25.00 & 34.95 & 82.86 & 84.58 & 54.43 & 57.66 & 61.76 & 64.42 & \textbf{90.97} & 76.30 & 62.20 & 59.55 & 65.43 & 64.29 \\
				MMGCN & 32.64 & 39.66 & 72.65 & 76.89 & 65.10 & 62.81 & 73.53 & 71.43 & 77.93 & 75.40 & 65.09 & 63.43 & 66.62 & 66.25 \\
				DialogueTRM & 61.11 & 57.89 & 84.90 & 81.25 & 69.27 & 68.56 & \textbf{76.47} & \textbf{75.99} & 76.25 & 76.13 & 53.39 & 58.09 & 68.51 & 68.20 \\
				MM-DFN & 44.44 & 44.44 & 77.55 & 80.00 & 71.35 & 66.99 & 75.88 & 70.88 & 74.25 & 76.42 & 58.27 & 61.67 & 67.84 & 67.85 \\
				TS-GCL  & \textbf{71.20} & 70.00 & 81.30 & 81.70 & 67.40 & 64.20 & 60.50 & 61.40 & 74.60 & 76.50 & 62.00 & 64.60 & 70.30 & 70.20 \\
				DialogueGCN & 64.29 & 29.03 & 80.86 & 64.37 & 43.14 & 50.96 & 68.49 & 63.29 & 71.85 & 68.19 & 57.68 & 62.41 & 62.07 & 58.19 \\
				GraphCFC* & 41.52 & 45.08 & \textbf{87.12} & \textbf{84.94} & 65.19 & 63.27 & 68.31 & 70.82 & 77.16 & 75.85 & 62.86 & 63.19 & 68.39 & 68.02 \\
				MultiEMO* & 53.80 & 56.29 & 83.33 & 83.50 & \textbf{75.60} &70.11 & 68.29 & 67.07 & 79.70 & 7579 & 64.82 & \textbf{70.35} & 72.29 & 71.69  \\
				SDT*  & 55.06 & 57.62 & 80.58 & 80.08 & 65.73 & 69.14 & 67.88 & 66.87 & 82.50 & 73.47 & 66.58 & 67.53 & 70.54 & 69.95 \\
				DEDNet* & 56.32 & \textbf{64.07} & 81.15 & 80.98 & 73.92 & \textbf{74.97} & 67.37 & 71.11 & 84.38 & 77.84 & 72.99 & 69.68 & 74.47 & 73.79 \\
				\midrule
				\textbf{Ours} & 59.35 & 61.54 & 84.08 & 84.08 & 70.91 & 73.75 & 69.73  & 72.68 & 84.73 & \textbf{81.18} & \textbf{74.64} & 67.98 & \textbf{75.17} & \textbf{74.87} \\
				\bottomrule
			\end{tabular}
		}
	\end{table*}

	\begin{table*}[!t]
		\centering
		\renewcommand{\arraystretch}{1.2} 
		\caption{Results on the MELD dataset.}
		\label{tab:meld_results}
		\resizebox{\linewidth}{!}{ 
			\begin{tabular}{l|*{14}{c}|cc}
				\toprule
				\multirow{4}{*}{Models} & \multicolumn{14}{c|}{MELD} & \multirow{4}{*}{w-Acc} & \multirow{4}{*}{w-F1} \\
				\cmidrule(lr){2-15}
				& \multicolumn{2}{c}{neutral} & \multicolumn{2}{c}{surprise} & \multicolumn{2}{c}{fear} & \multicolumn{2}{c}{sadness} & \multicolumn{2}{c}{joy} & \multicolumn{2}{c}{disgust} & \multicolumn{2}{c|}{anger} &  &  \\
				\cmidrule(lr){2-3}\cmidrule(lr){4-5}\cmidrule(lr){6-7}\cmidrule(lr){8-9}\cmidrule(lr){10-11}\cmidrule(lr){12-13}\cmidrule(lr){14-15}
				& ACC & F1 & ACC & F1 & ACC & F1 & ACC & F1 & ACC & F1 & ACC & F1 & ACC & F1 &  &  \\
				\midrule
				DialogueRNN & 82.17 & 76.56 & 46.62 & 47.64 & 0.00 & 0.00 & 21.15 & 24.65 & 49.50 & 51.49 & 0.00 & 0.00 & 48.41 & 46.01 & 60.27 & 57.95 \\
				MMGCN & \textbf{84.32} & 76.96 & 47.33 & 49.63 & 2.00 & 3.64 & 14.90 & 20.39 & 56.97 & 53.76 & 1.47 & 2.82 & 42.61 & 45.23 & 61.34 & 58.41 \\
				DialogueTRM & 83.20 & 79.41 & 56.94 & 55.27 & 12.00 & 17.39 & 27.88 & 36.48 & 60.45 & 60.30 & 16.18 & 20.18 & 51.01 & 49.79 & 65.10 & 63.80 \\
				MM-DFN & 79.06 & 75.80 & 53.02 & 50.42 & 0.00 & 0.00 & 17.79 & 23.72 & 59.20 & 55.48 & 0.00 & 0.00 & 50.43 & 48.27 & 60.96 & 58.72 \\
				TS-GCL &  78.10 & \textbf{80.60} & 56.70 & 56.40 & 6.80 & 5.20 & 42.30 & 43.70 & \textbf{68.30} & \textbf{66.30} & 2.30 & 2.60 & 43.80 & 48.50 & 64.40 & 64.10  \\
				GraphCFC* & 71.26 & 75.17 & 46.18 & 45.68 & 9.09 & 3.28 & 30.43 & 11.02  & 50.26 & 49.49 & 0.00 & 0.00 & 35.88 & 41.93 & 54.34 & 55.20  \\
				MultiEMO* & 76.44 & 79.42 & 55.92 & 58.12 & 25.71 & 21.18 & 51.05 & 41.60 & 62.23 & 63.07 & \textbf{45.71} & 31.07 & \textbf{55.28} & 53.37 & 65.45 & 65.77 \\
				SDT* & 75.99 & 79.65 & 59.21 & 58.78 & 24.44 & \textbf{23.16} & \textbf{59.22} & 39.23 & 64.40 & 62.76 & 40.00 & \textbf{31.86} & 52.27 & 54.44 & 66.00 & 65.92 \\
				DEDNet* & 76.51 & 79.97 & 56.77 & 58.90 & 21.88 & 17.07 & 50.38 & 39.30 & 64.97 & 62.63 & 40.54 & 28.57 & 53.65 & \textbf{54.49} & 65.52 & 65.88 \\
				\midrule
				\textbf{Ours} & 77.59 & 79.92 & \textbf{59.43} & \textbf{59.43} & \textbf{26.09} & 16.44 & 49.10 & \textbf{43.73} & 65.31 & 64.48 & 40.00 & 21.51 & 51.29 & 54.30 &  \textbf{66.52} & \textbf{66.62} \\
				\bottomrule
			\end{tabular}
		}
	\end{table*}

	\subsection{Modality Importance Analysis}
	
	As shown in Table \ref{tab:modal_performance}, the comparison of emotion recognition performance among unimodal, bimodal, and trimodal systems indicates that the contribution of the text modality is significantly higher than that of the audio and visual modalities. This not only reflects the susceptibility of the latter two to noise interference but also highlights the dominant role of the text modality, thereby validating the effectiveness of the proposed fusion module.
	
	\begin{table}[!t]
		\centering
		\caption{Performance under different modal settings across datasets.}
		\label{tab:modal_performance}
		\resizebox{\linewidth}{!}{
			\begin{threeparttable}
				\setlength{\tabcolsep}{12pt}{
					\begin{tabular}{c|cc|cc}
						\hline
						\multirow{2}{*}{Modality Setting} & \multicolumn{2}{c|}{IEMOCAP} & \multicolumn{2}{c}{MELD} \\ \cline{2-5} 
						& w-Acc & w-F1 & w-Acc & w-F1 \\ \hline
						V   & 40.11 & 31.25 & 23.16 & 31.27 \\ 
						A   & 54.75 & 51.17 & 36.15 & 39.01 \\ 
						T   & 69.86 & 69.34 & 65.05 & 65.11 \\ 
						V+T & 69.52 & 68.92 & 65.81 & 65.93 \\
						A+T & 74.05 & 72.98 & 65.67 & 65.8 \\ 
						\hline
						A+V+T & \textbf{75.17} & \textbf{74.87} & \textbf{66.52} & \textbf{66.62} \\ \hline
				\end{tabular}}
				\begin{tablenotes}[para,flushleft]
					\footnotesize
					A, V, and T represent the acoustic, visual, and textual modalities, respectively.
				\end{tablenotes}
			\end{threeparttable}
		}
	\end{table}
	
	\subsection{Sliding Window Effect}
	In multimodal emotion recognition tasks, the window size determines the length of contextual information considered by the model when processing temporal data. A smaller window may fail to capture sufficient information, while a larger window may introduce noise and increase computational complexity. Therefore, selecting an appropriate window size helps to achieve a balance between information capture and computational efficiency.
	
\begin{figure}[!t]
		\centering
		\subfloat{
			\includegraphics[width=0.49\linewidth]{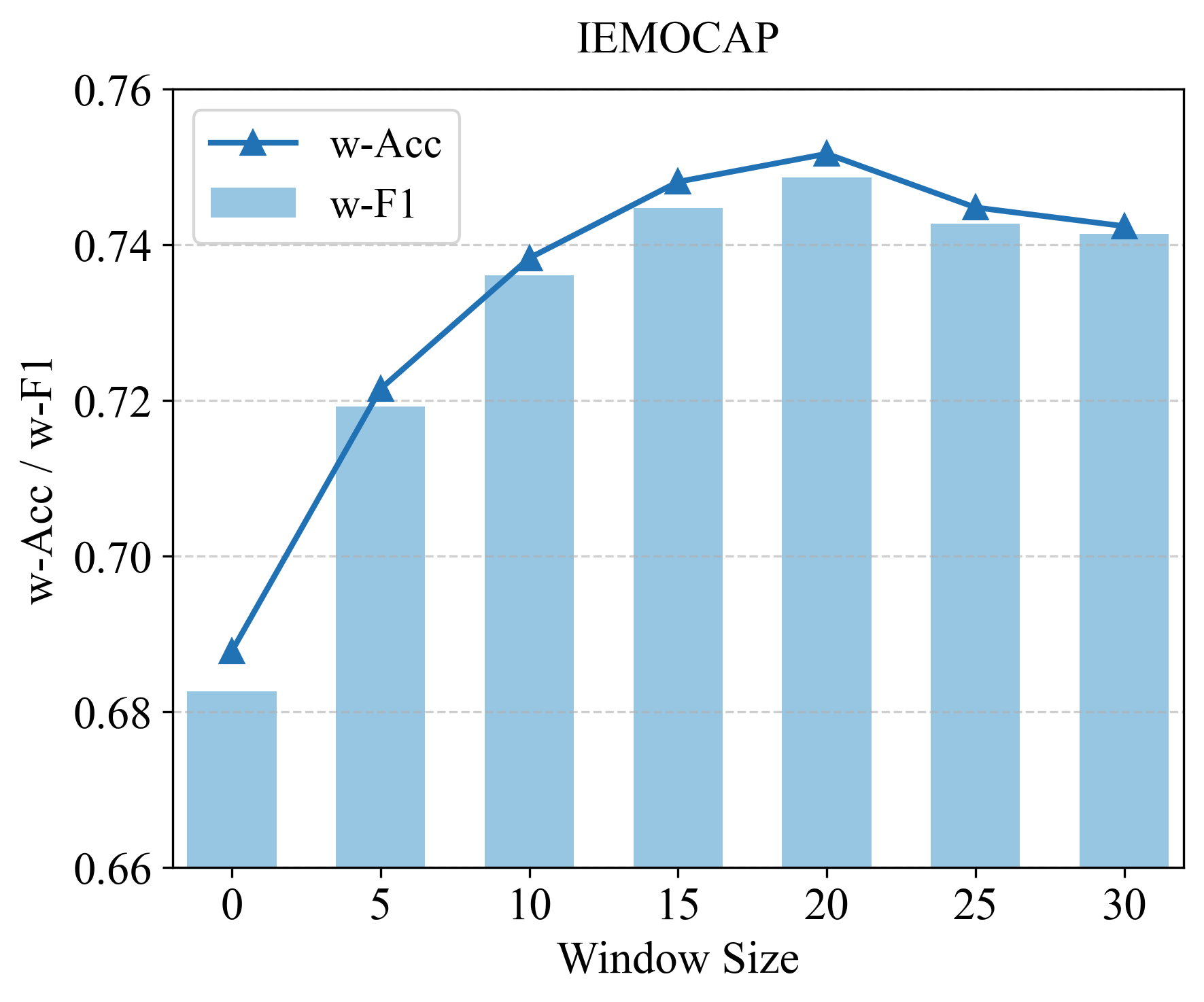}
		}
		\subfloat{
			\includegraphics[width=0.49\linewidth]{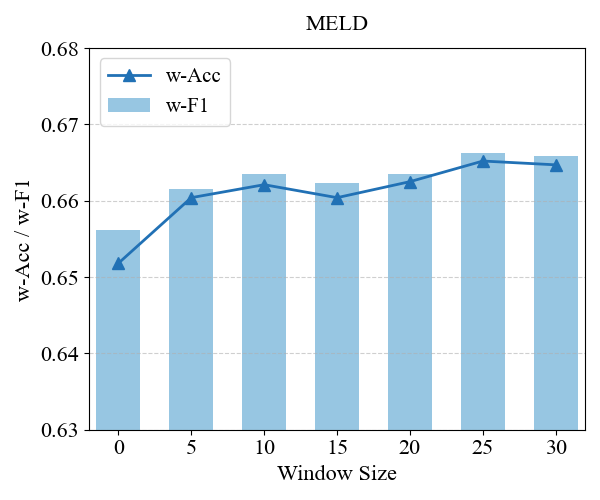}
		}
		\caption{Effect of varying window sizes on model performance across two datasets.}
		\label{fig:window}
	\end{figure}

	As shown in Figure \ref{fig:window}, we varied the window size from 0 to 30 in our experiments and observed that the optimal window size is \textbf{20} for the IEMOCAP dataset and \textbf{25} for the MELD dataset. This may be attributed to the nature of the datasets: utterances in IEMOCAP tend to be longer with slower emotional transitions, making a larger window beneficial for capturing complete contextual cues. In contrast, MELD features faster-paced conversations and more frequent emotional shifts, where a larger window helps the model to capture emotional dynamics across multiple utterances. These findings indicate that the optimal window size is closely tied to the structural characteristics of each dataset. Moreover, using a fixed sliding window yields approximately linear inference complexity with respect to dialogue length, and the performance gains generally saturate beyond a moderate window size, which is favorable for scaling to longer conversational streams.

	 In addition, we further evaluate the impact of inference-time temporal segmentation on performance. As shown in Table \ref{tab:temporal_segmentation_robustness}, under no segmentation, 50\% overlapping segmentation, and non-overlapping segmentation settings, the performance on both datasets exhibits a smooth degradation trend, indicating that the model maintains good stability under sequence segmentation. Meanwhile, overlapping segmentation performs slightly better than non-overlapping segmentation, suggesting that overlapping windows help mitigate the impact of boundary information fragmentation.
	
\begin{table}[!t]
	\centering
	\caption{ Performance under different temporal segmentation settings at inference time.}
	\label{tab:temporal_segmentation_robustness}
	\resizebox{\linewidth}{!}{%
		\begin{threeparttable}
			\setlength{\tabcolsep}{12pt}{
				\begin{tabular}{l|cc|cc}
					\hline
					\multirow{2}{*}{Temporal Segmentation Setting} & \multicolumn{2}{c|}{IEMOCAP} & \multicolumn{2}{c}{MELD} \\
					\cline{2-5}
					& w-Acc & w-F1 & w-Acc & w-F1 \\
					\hline
					No-overlap   & 72.86 & 72.21 & 66.22 & 66.35 \\
					50\% overlap & 72.92 & 72.28 & 66.41 & 66.55 \\
					Baseline     & \textbf{75.17} & \textbf{74.87} & \textbf{66.52} & \textbf{66.62} \\
					\hline
			\end{tabular}}
			\begin{tablenotes}[para,flushleft]
				\footnotesize
				Temporal segmentation is applied only at inference time. Baseline denotes no temporal segmentation.
			\end{tablenotes}
		\end{threeparttable}
	}
\end{table}
	
	Moreover, the relation-subgraph construction still has certain limitations. Our method explicitly distinguishes intra-speaker and inter-speaker relations, and therefore relies on reliable speaker annotations. The comparison results in Table \ref{tab:graph_robustness_speaker} show that when speaker identity is not used during graph construction or when speaker noise is introduced, performance on both datasets degrades smoothly. This indicates that speaker priors provide gains, while the method still retains a certain level of stability when speaker information is imperfect. For more complex conversational settings, generalization may be further improved by incorporating pseudo-speaker annotations, lightweight edge gating, and sparse long-range connections.
	
	\begin{table}[!t]
		\centering
		\caption{Robustness of relation-subgraph construction under missing or noisy speaker information (w-F1).}
		\label{tab:graph_robustness_speaker}
		\begin{threeparttable}
			\resizebox{\linewidth}{!}{%
				\setlength{\tabcolsep}{12pt}{
					\begin{tabular}{lccc}
						\hline
						Dataset & No-speaker & Noisy-speaker ($p=0.3$) & Ours \\
						\hline
						IEMOCAP & 70.02 & 71.19 & \textbf{74.87} \\
						MELD    & 66.15 & 66.32 & \textbf{66.62} \\
						\hline
					\end{tabular}%
				}
			}
		\end{threeparttable}
	\end{table}
	
	\subsection{Self-Supervised Loss Effect}
	As shown in Table \ref{tab:MSL}, the introduction of the modality self-supervised loss significantly improves the model’s performance. For each unimodal loss, the weight factor \(\lambda\) is set to one-tenth of its corresponding loss value. This strategy dynamically adjusts the weight factors, enabling the model to focus more on tasks with larger losses, thereby enhancing the performance of weaker modalities while maintaining the dominance of the multimodal fusion loss.
	
	\begin{table}[!t]
		\centering
		\caption{Performance of the model under different loss strategies.}
		\label{tab:MSL}
		\begin{threeparttable}
			\resizebox{\linewidth}{!}{
            \setlength{\tabcolsep}{12pt}{
            \begin{tabular}{c c c|c c}
					\hline
					Interaction & \multicolumn{2}{c|}{IEMOCAP} & \multicolumn{2}{c}{MELD} \\
					Strategy & w-Acc        & w-F1        & w-Acc        & w-F1        \\
					\hline
					w/o MSL              & 72.58 & 72.52 & 65.78 & 66.05 \\
					w MSL                & \textbf{75.17} & \textbf{74.87} & \textbf{66.52} & \textbf{66.62} \\
					\hline
			\end{tabular}}}
			\begin{tablenotes}[para,flushleft]
				\footnotesize
				MSL represents Modality-Specific Loss.
			\end{tablenotes}
		\end{threeparttable}
	\end{table}

\begin{figure}[!t]
		\centering
		\includegraphics[width=\linewidth]{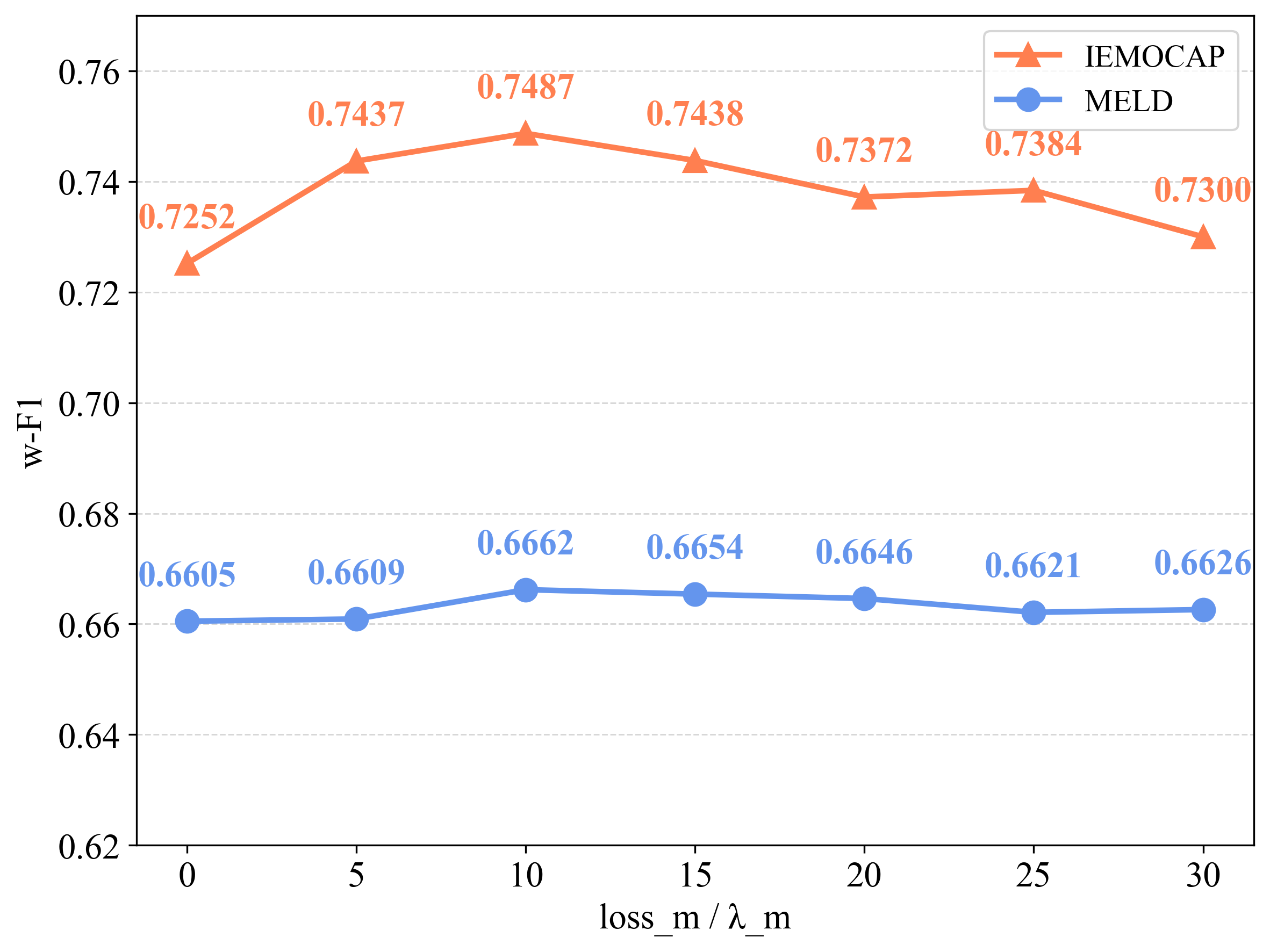}
		\caption{The impact of different $\lambda_m$ values on the model performance. Note that a horizontal coordinate of 0 means that only fusion loss is used as the total loss.}
		\label{fig:损失参数设定}
	\end{figure}

	We set the weight factor $\lambda_m$ for the modality-specific losses to range from $1/5$ to $1/30$, in order to balance unimodal and fusion effects. As shown in  Figure \ref{fig:损失参数设定}, when $\lambda_m = 1/10$, the model achieves the best performance on both the MELD and IEMOCAP datasets, demonstrating the model's adaptability to different datasets.

	\subsection{Denoising Effect Analysis}
	
\begin{figure}[!t]
		\centering
		\subfloat{
			\includegraphics[width=0.49\linewidth]{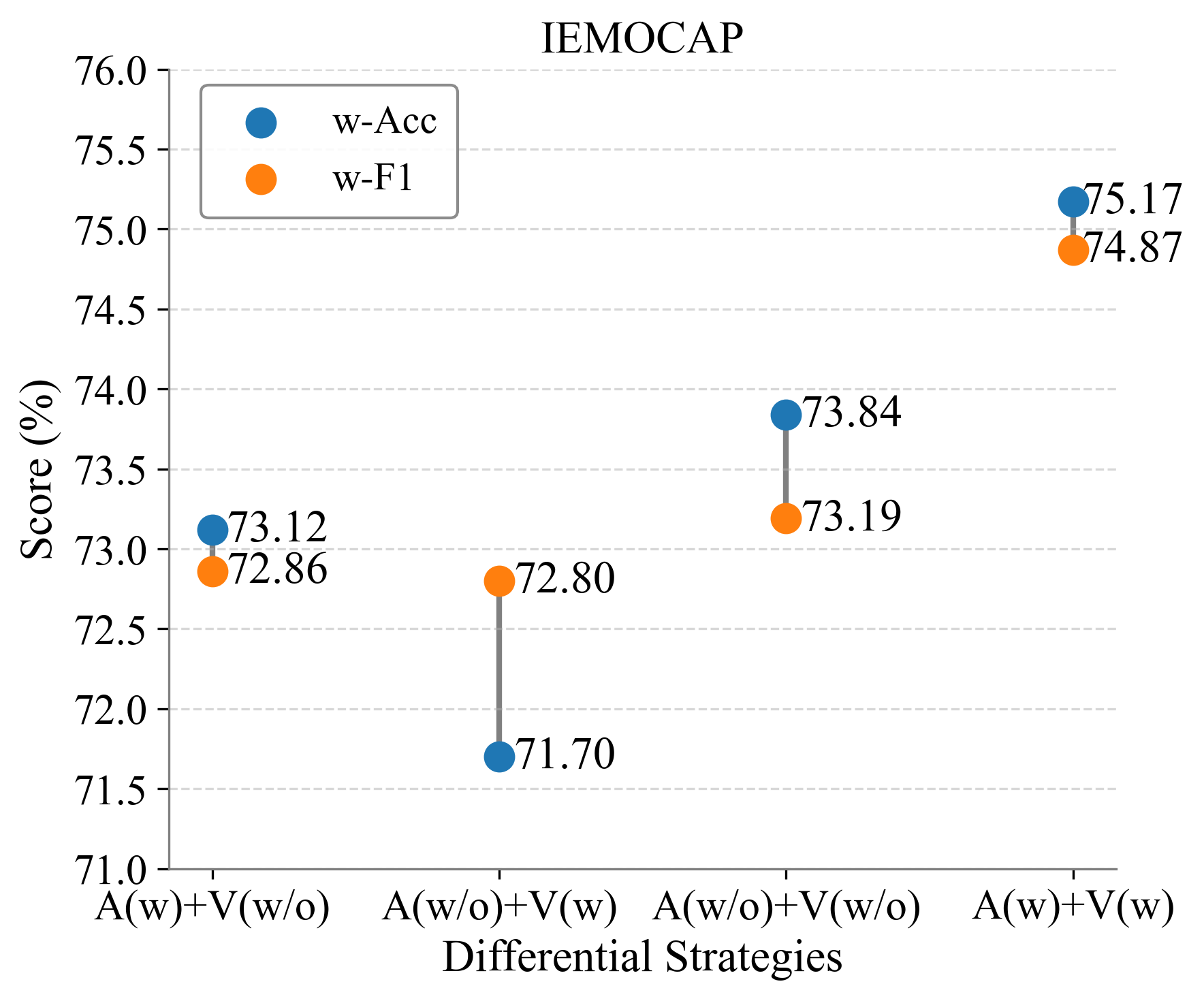}
			\label{IEMOCAP情感转移样本}
		}
		\subfloat{
			\includegraphics[width=0.49\linewidth]{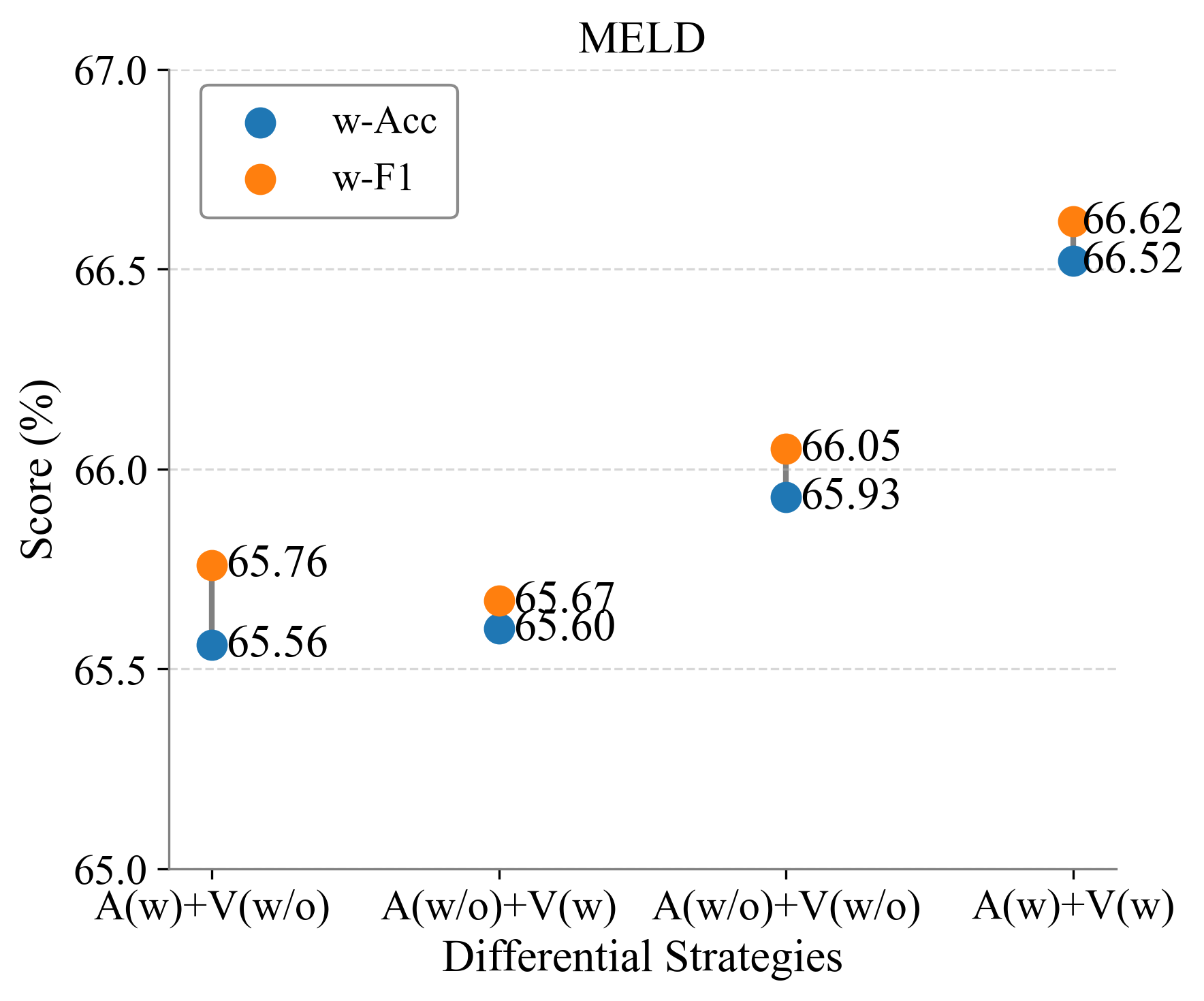}
			\label{fig:MELD情感转移样本}
		}
		\caption{Performance comparison of differential strategies on the IEMOCAP and MELD datasets.}
		\label{fig:差分效果}
	\end{figure}

	We evaluated the effectiveness of the Differential Transformer in denoising and performance enhancement for audio and visual modalities. As shown in Figure \ref{fig:差分效果}, when both modalities employ the Differential Transformer, the model accurately captures dynamic emotional cues and effectively reduces noise, significantly improving recognition performance. When neither modality uses the Differential Transformer, the model maintains good performance by relying on static feature consistency. However, applying the Differential Transformer to only one modality disrupts feature homogeneity and lacks collaborative validation, resulting in poor denoising and potential interference, with performance worse than when it is not used. These results demonstrate the effectiveness of the Differential Transformer in denoising and collaborative modeling of multimodal dynamic features.
	
	\subsection{Subgraph Interaction Analysis}
\begin{figure}[!t]
		\centering
		\subfloat{
			\includegraphics[width=0.5\linewidth]{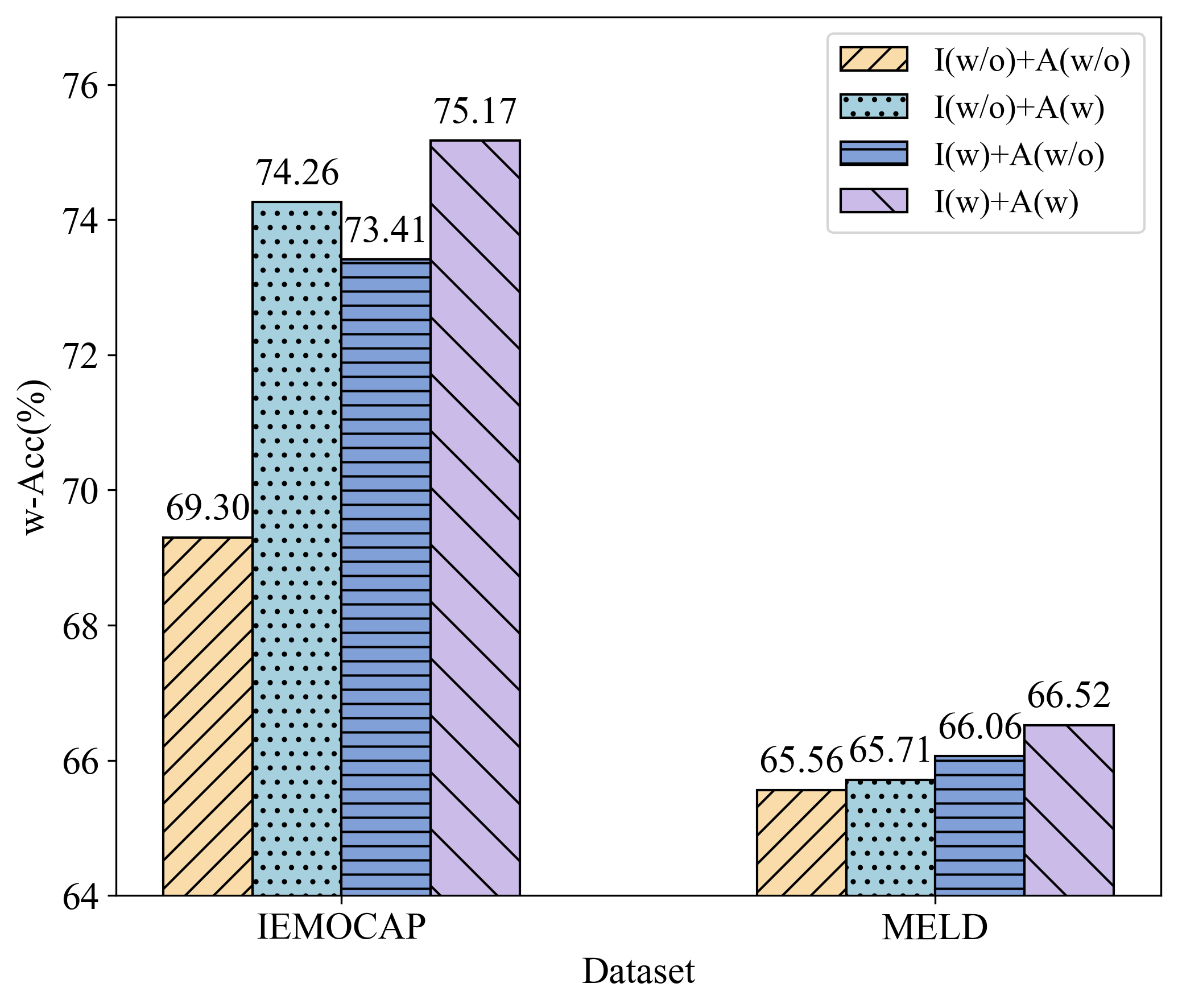}
		}
		\subfloat{
			\includegraphics[width=0.5\linewidth]{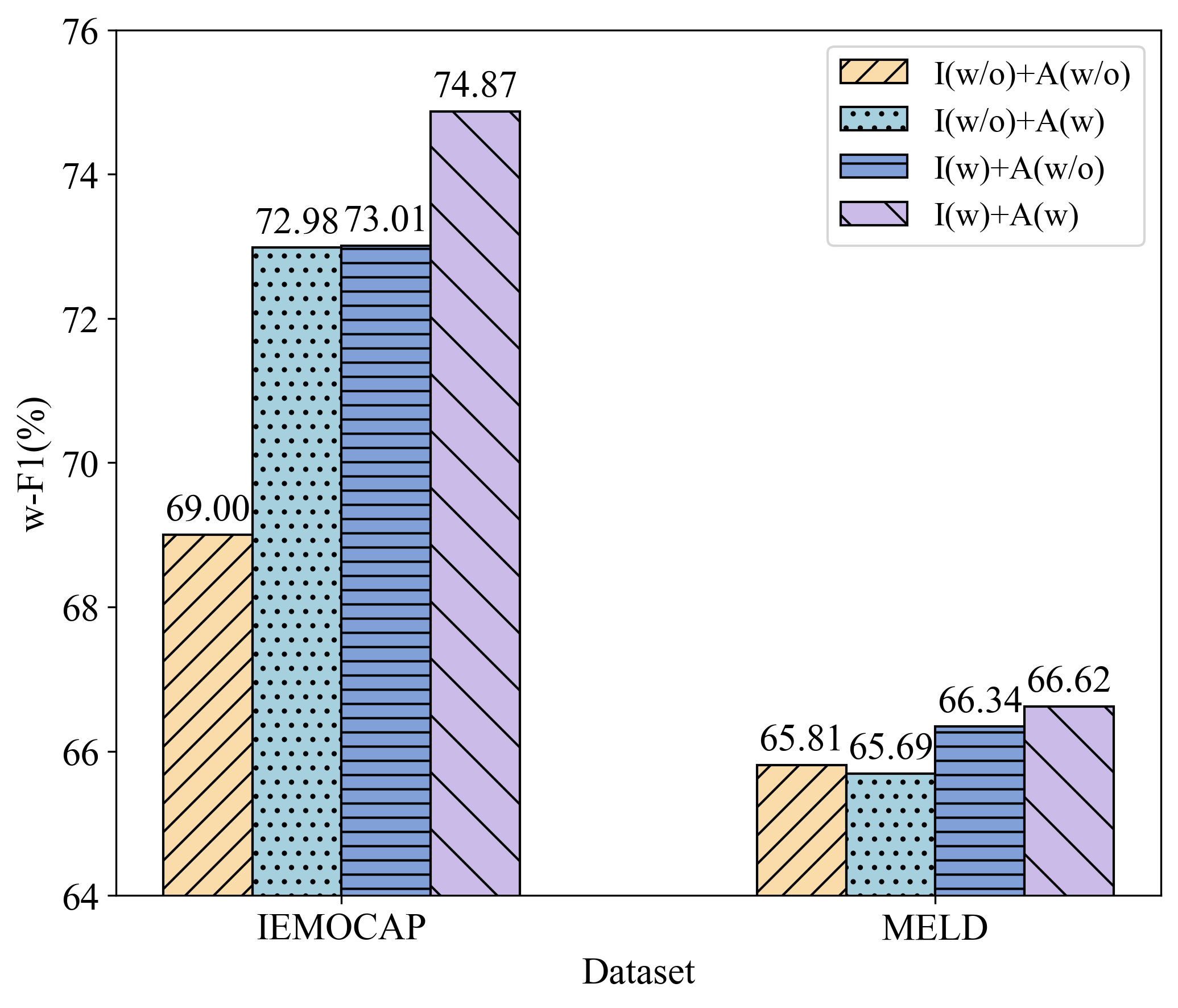}
		}
		\caption{Effect of relational subgraph modules on the overall model performance (I denotes InterGAT, A denotes IntraGAT).}
		\label{fig:图有无}
	\end{figure}

	To validate the performance enhancement of the designed InterGAT and IntraGAT, ablation experiments were conducted. As shown in Figure \ref{fig:图有无}, the model performs optimally on both datasets when the dual graph structures are used in parallel (w, w). While using a single graph structure also improves performance, the complementary nature of the dual graph structures allows for more comprehensive exploration of the textual feature relationships. This confirms the necessity of the dual graph structure design to enhance the model's emotional recognition capability and provides key support for the effectiveness of the model’s feature interaction mechanism.
	
	To verify the impact of the interaction order between the cross-speaker graph and the speaker-specific graph on feature extraction, we compared four strategies: Single Graph, Parallel Interaction, Relational Incremental Interaction, and Incremental Interaction. As shown in Table \ref{tab:interaction_strategy_performance}, the Single Graph performed poorly on the IEMOCAP dataset, likely due to the loss of hierarchical features. However, on the MELD dataset, its performance was comparable to some parallel interaction strategies, indicating that the advantage of the dual graph structure depends on dataset characteristics. Parallel and relational incremental interactions showed similar performance, while incremental interaction achieved the best results on both datasets, confirming the rationality of the hierarchical design that first models global associations across speakers and then extracts individual features.

	\begin{table}[!t]
		\centering
		\caption{Performance under different interaction strategies evaluated across datasets.}
		\label{tab:interaction_strategy_performance}
		\renewcommand{\arraystretch}{1.2} 
		\resizebox{\linewidth}{!}{ 
			\begin{threeparttable}
            \setlength{\tabcolsep}{12pt}{
				\begin{tabular}{l|p{2.8cm}|cc|cc} 
					\hline
					\multicolumn{2}{c|}{\multirow{2}{*}{\textbf{Strategy}}} & \multicolumn{2}{c|}{\textbf{IEMOCAP}} & \multicolumn{2}{c}{\textbf{MELD}} \\ \cline{3-6} 
					\multicolumn{2}{c|}{} & \textbf{w-Acc} & \textbf{w-F1} & \textbf{w-Acc} & \textbf{w-F1} \\ \hline
					\multicolumn{2}{c|}{SingleGraph} & 70.98 & 70.64 & 66.13 & 66.17 \\ \hline
					\multirow{4}{*}{SubGraphs} & PIS (Sum) & 74.04 & 73.52 & 65.65 & 65.97 \\ 
					& PIC (Concat) & 73.27 & 72.81 & 65.97 & 66.12 \\ 
					& Relncremental & 74.35 & 73.77 & 65.87 & 66.19 \\ 
					& Incremental & \textbf{75.17} & \textbf{74.87} & \textbf{66.52} & \textbf{66.62} \\ \hline
				\end{tabular}}
				\begin{tablenotes}[para,flushleft]
					\footnotesize
					SingleGraph: Merge subgraphs into one graph. 
					SubGraphs strategies: 
					PIS = Parallel Interaction+Sum; 
					PIC = Parallel Interaction+Concat; 
					Relncremental = Relncremental Interaction; 
					Incremental = Incremental Interaction.
				\end{tablenotes}
			\end{threeparttable}
		}
	\end{table}

	
	\subsection{Emotion Dependency Analysis}
	Dynamic emotion dependency recognition aims to uncover emotional dependency relationships induced by speaker interactions in consecutive utterances, which is a critical issue in sentiment analysis. To validate the performance advantages of our proposed model on this task, we conducted comparative experiments against four representative baseline models: MM-DFN, SDT, MultiEMO and GraphCFC.
	
	Regarding sample definition, consecutive utterances with the same emotion label were categorized as inter-speaker dependency samples if uttered by different speakers, and as intra-speaker dependency samples if uttered by the same speaker. All dependency samples were extracted from the original test set to construct a dedicated evaluation set, ensuring the training set remained unchanged and focusing the assessment on test performance.
	
\begin{figure*}[!t]
		\centering
		\includegraphics[width=\linewidth]{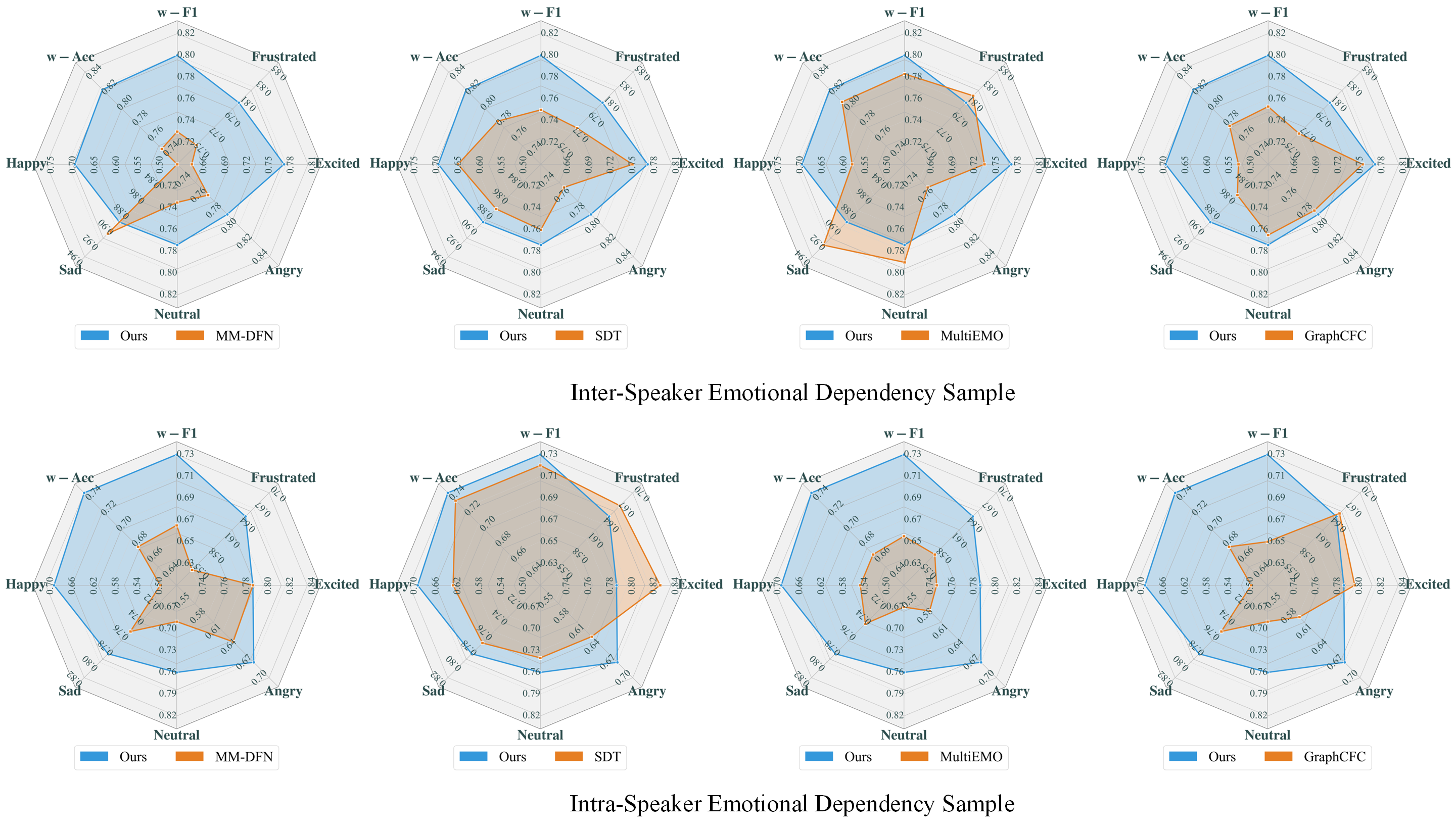}
		\caption{The figure illustrates the performance comparison between our model and five baseline models on inter-speaker emotion dependency samples and intra-speaker emotion dependency samples in the IEMOCAP testing set.}
		\label{fig:雷达图}
	\end{figure*}

	As shown in Figure \ref{fig:雷达图}, compared to the baseline models, our model demonstrates a larger coverage area across all metrics, which fully attests to the efficacy of both the Differential Denoising Module and the Text-Guided Cross-Modal Diffusion Attention Fusion Module. Specifically, the Differential Denoising Module effectively suppresses noise interference in the audio and visual modalities, enhancing the purity of dynamic features and thus improving the model’s sensitivity to emotional changes and recognition accuracy. Meanwhile, the text-guided cross-modal diffusion attention fusion module strengthens the guidance of textual information over multimodal features, enabling deep interaction and complementary enhancement among modalities, thereby promoting consistency and complementarity of multimodal features. The synergistic effect of these two modules allows the model to more accurately capture emotional dependencies in complex dialogues, significantly boosting the performance and robustness of dynamic emotion dependency recognition.
	
	\subsection{Emotion Shift Analysis}
	
\begin{figure}[!t]
		\centering
		\subfloat{
			\includegraphics[width=0.5\linewidth]{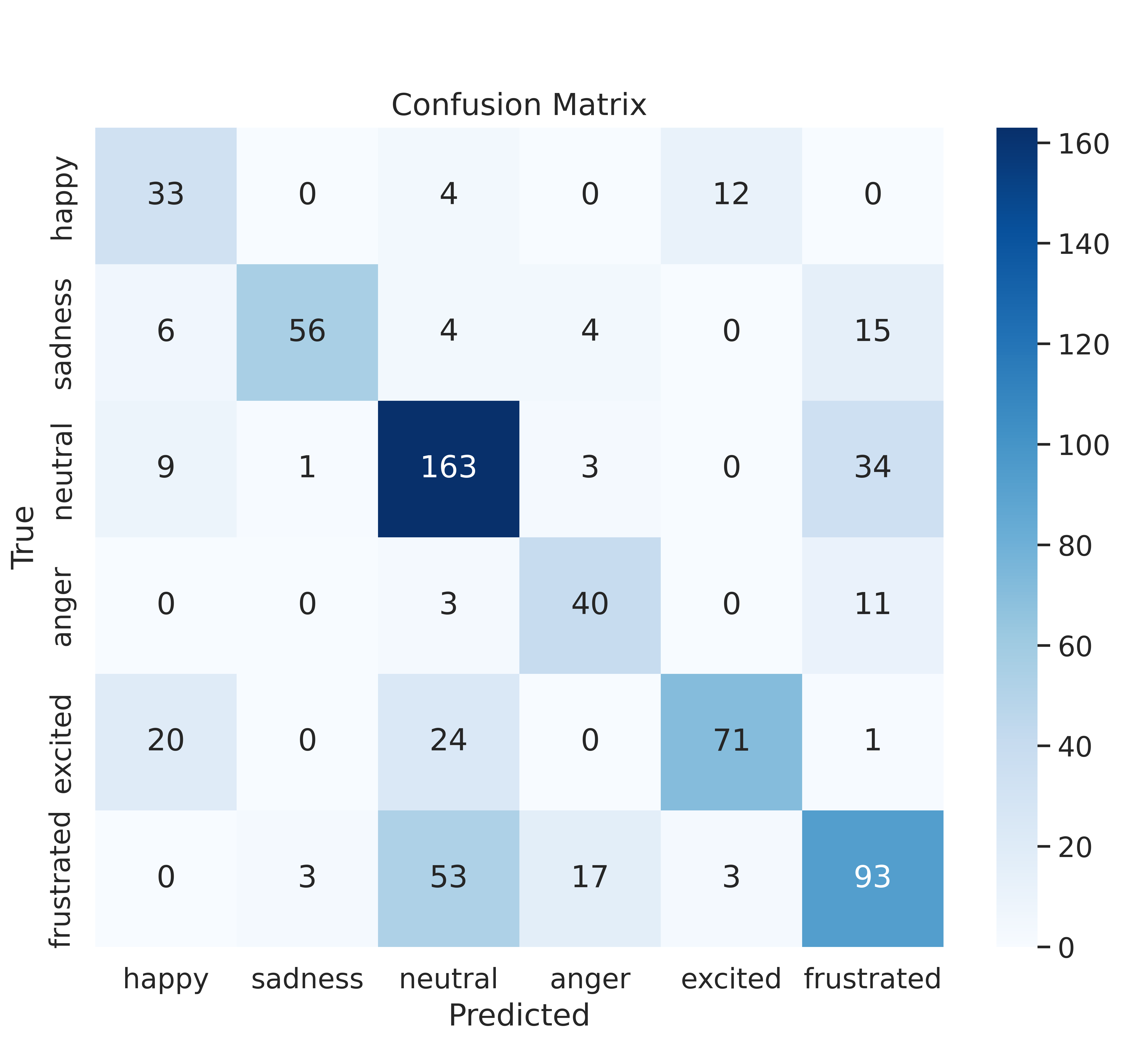}
		}
		\subfloat{
			\includegraphics[width=0.5\linewidth]{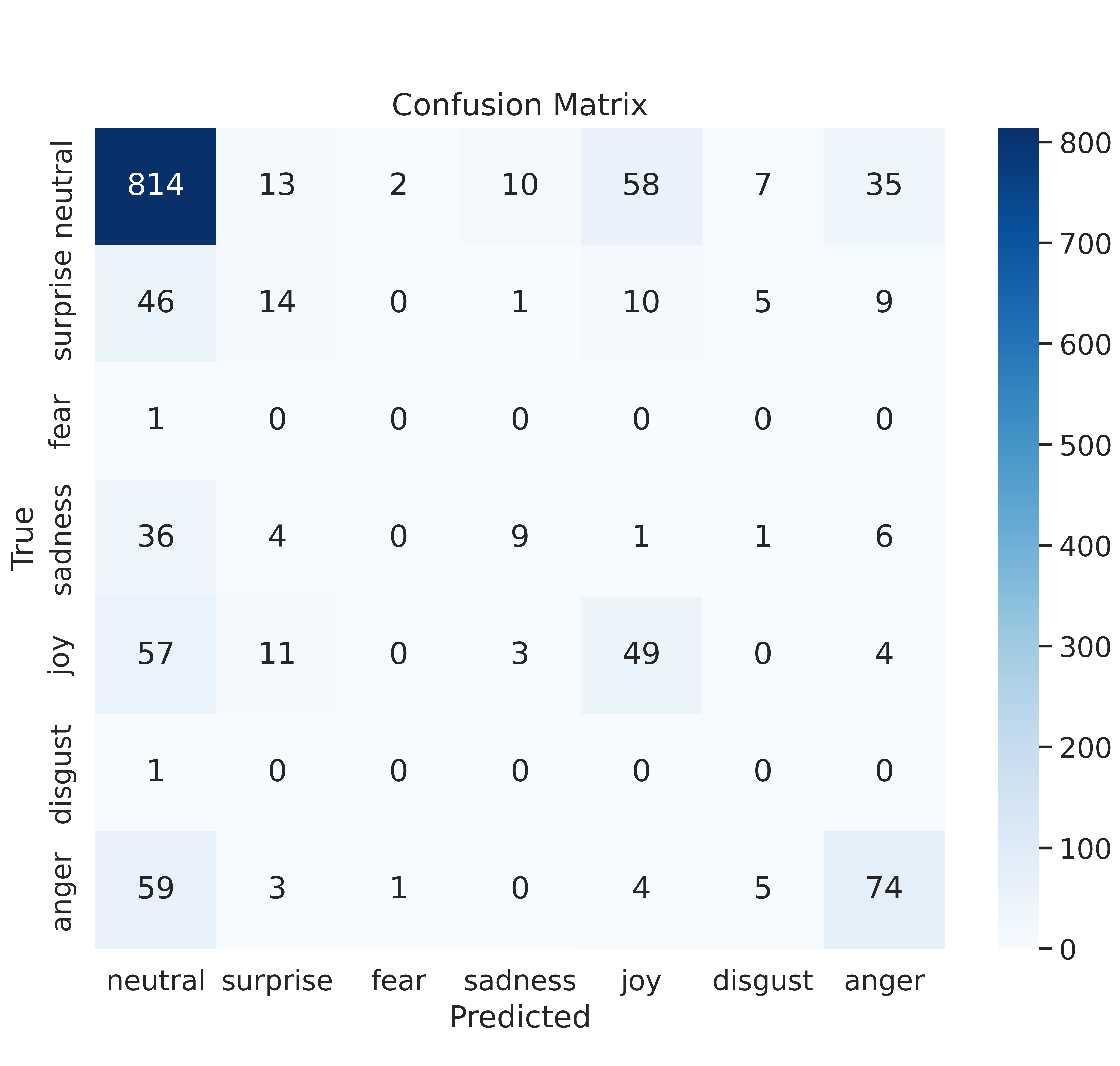}
		}
		\caption{\textcolor{black}{The figure shows the confusion matrices for our model evaluated on emotion shift samples from the IEMOCAP and MELD datasets, with rows representing the actual labels and columns representing the predicted labels.}}
		\label{fig:情感转移}
	\end{figure}

	Identifying dialogues with emotional shifts is a significant challenge in sentiment analysis, requiring the model not only to accurately recognize emotion categories but also to effectively capture dynamic emotional changes. To evaluate the performance of the proposed model on this task, we extracted samples containing emotional shifts from the test set of the original dataset, keeping the training set unchanged and focusing on this specific test set. As shown in Table \ref{tab: Emotion_Shif} and Figure \ref{fig:情感转移}, the model achieves significant improvements across all metrics.
	
	The performance gains mainly result from the synergistic effect of two core modules. The differential denoising module effectively removes noise from audio and visual data, highlighting key dynamic features at moments of emotional shift, making the signals clearer. Meanwhile, the text-guided cross-modal diffusion attention fusion module strengthens the correlation between audio, visual, and textual features, enabling more accurate identification of the timing and intensity of emotional shifts. Together, these modules improve the model’s sensitivity and discriminative ability to emotional changes, significantly enhancing recognition accuracy and robustness.
	
	\begin{table}[!t]
		\centering
		\caption{Performance of the model on emotion shift samples.}
		\label{tab: Emotion_Shif}
		\resizebox{\linewidth}{!}{
        \setlength{\tabcolsep}{12pt}{
        \begin{tabular}{l|cc|cc} 
				\hline
				\multirow{2}{*}{Models} & \multicolumn{2}{c|}{IEMOCAP} & \multicolumn{2}{c}{MELD} \\ \cline{2-5} 
				& w-Acc & w-F1 & w-Acc & w-F1 \\ \hline
				MM-DFN & 53.77 & 53.34 & 59.62 & 58.19 \\ 
				SDT & 53.41 & 52.9 & 60.61 & 58.67 \\ \hline
				\textbf{Ours} & \textbf{63.93} & \textbf{64.05} & \textbf{63.02} & \textbf{63.21} \\ \hline
		\end{tabular}}}
		\label{tab:model_performance}
	\end{table}

	\subsection{Computational Efficiency Analysis}
	
	To evaluate the computational cost and runtime efficiency of our model, we conduct analyses from both theoretical complexity and empirical measurements. For a standard Transformer self-attention layer, the dominant computations consist of linear projections and attention operations: the $\mathbf{Q}/\mathbf{K}/\mathbf{V}$ projections have complexity $O(BLd^2)$, and the attention scoring with weighted summation has complexity $O(BL^2d)$, where $B$ denotes the batch size, $L$ the dialogue window length, and $d$ the hidden dimension. Our differential attention preserves the same leading term $O(BL^2d)$, while introducing an additional time-referenced attention map and a differencing operation; this incurs only constant-factor overhead and does not change the asymptotic order. Under a fixed number of layers and attention heads, the relation-graph branch scales linearly with the number of edges and can be written as $O(B|E|d)$. The additional cost of the fusion branch mainly arises from cross-modal attention and gating computations, which are likewise implemented by a small number of matrix operations. Since we adopt a fixed sliding-window setting (IEMOCAP: $L=20$; MELD: $L=25$), these overheads remain well controlled within the windowed modeling framework. 
	
	Empirically, we report the parameter size and inference time on the full test set (forward pass only, seconds per full test set) and compare them with representative methods (see Table~\ref{tab:efficiency}). We further report the peak GPU memory usage during inference to characterize the upper bound of resource consumption: approximately 57.6~MB on MELD  and 259.3~MB on IEMOCAP. Overall, while achieving leading performance, our method maintains a controllable parameter budget and demonstrates competitive inference efficiency and memory footprint on both datasets.
\begin{table}[!t]
	\centering
	\caption{Model complexity and inference time comparison on IEMOCAP and MELD.}
	\label{tab:efficiency}
	\resizebox{\linewidth}{!}{ 
		\begin{threeparttable}
			\setlength{\tabcolsep}{12pt}{
				\begin{tabular}{c|c|c|c}
					\hline
					\multirow{2}{*}{Model} & \multirow{2}{*}{\# Params.} & \multicolumn{1}{c|}{IEMOCAP} & \multicolumn{1}{c}{MELD} \\
					\cline{3-4}
					& & Inference Time & Inference Time \\
					\hline
					MMGCN   & 5.63M  & 53.7s  & 75.3s  \\
					RGAT    & 15.28M & 68.5s  & 146.3s \\
					DER-GCN & 78.59M & 125.5s & 189.7s \\
					M3Net   & 7.88M  & 57.0s  & 87.6s  \\
					\textbf{Ours} & \textbf{11.48M} & \textbf{61.2s} & \textbf{98.5s} \\
					\hline
			\end{tabular}}
			\begin{tablenotes}[para,flushleft]
				\footnotesize
				Inference time denotes the forward-pass-only runtime on the full test set (in seconds). All timing results were measured under the same server environment and timing protocol (NVIDIA A800 80GB PCIe GPU, AMD EPYC 7763 64-Core Processor, and approximately 503 GB RAM; CUDA 12.4) with a unified inference batch size of 16 for both IEMOCAP and MELD.
			\end{tablenotes}
		\end{threeparttable}
	}
\end{table}

	\subsection{Ablation Study}
	This section quantitatively evaluates the effectiveness of the key design choices from three perspectives: component contributions, sensitivity to the cross-modal diffusion fusion coefficient, and comparison with conventional denoising strategies.
	
	\textbf{Component-wise Ablation Study}
	To quantify the independent contribution of each module, we conduct ablations by removing the differential, gating, relation-graph, and diffusion-fusion components one at a time while keeping the training settings and evaluation protocol unchanged. The results are reported in Table~\ref{tab:ablation_all}. Removing any component leads to performance degradation, indicating that the overall improvement arises from complementary cooperation among modules.
	
	Removing the differential module consistently degrades performance on both datasets, suggesting that the first-order differencing in the relation domain effectively suppresses relation redundancy that remains consistent across adjacent time steps. Removing the gating module also causes a consistent drop, demonstrating that the gate performs context-aware filtering over the differential responses after residual fusion and further mitigates random jumps induced by abrupt perturbations. The largest decrease is observed when removing the relation-graph module, highlighting the importance of structured interactions for modeling emotional dependencies. Removing the diffusion fusion module also yields notable degradation, confirming its independent gains in alleviating modality imbalance and facilitating information transfer.
	
	\begin{table}[!t]
		\centering
		\caption{Results of component-wise ablation study.}
		\label{tab:ablation_all}
		\begin{threeparttable}
			\resizebox{\linewidth}{!}{%
            \setlength{\tabcolsep}{12pt}{
				\begin{tabular}{lcc|cc}
					\toprule
					\multirow{2}{*}{Setting} & \multicolumn{2}{c|}{IEMOCAP} & \multicolumn{2}{c}{MELD} \\
					& w-Acc & w-F1 & w-Acc & w-F1 \\
					\midrule
					w/o Differential & 73.84 & 73.19 & 65.93 & 66.05 \\
					w/o Gating       & 74.01 & 74.83 & 66.15 & 66.33 \\
					w/o Graph        & 69.30 & 69.00 & 65.56 & 65.81 \\
					w/o Diffusion    & 72.92 & 72.41 & 65.88 & 66.08 \\
					\textbf{Full Model}       & \textbf{75.17} & \textbf{74.87} & \textbf{66.52} & \textbf{66.62} \\
					\bottomrule
				\end{tabular}%
                }
			}
		\end{threeparttable}
	\end{table}
	
	\textbf{Sensitivity to the Diffusion Fusion Coefficient}
	We further analyze the sensitivity to the coefficient $\gamma \in [0,1]$ in the cross-modal diffusion-attention fusion, as summarized in Table~\ref{tab:gamma_sensitivity}. According to  \eqref{eq:fusion_weight}, $\gamma$ balances the contribution between the normalized diffusion correlation term $(\hat{\mathbf{S}}_{t}\hat{\mathbf{S}}_{m}^{\top})$ and the original correlation term $\mathbf{J}_{m}$. When $\gamma=0$, the fusion degenerates to using only $\mathbf{J}_{m}$; when $\gamma=1$, only the diffusion correlation term is retained. The results show an overall improvement as $\gamma$ increases, with performance saturating in the mid-to-high range. This suggests that the diffusion correlation term is crucial for stable cross-modal correlation estimation, while the model remains reasonably robust over a wide range of $\gamma$.
	
	\begin{table}[!t]
		\centering
		\caption{Sensitivity analysis of the diffusion fusion coefficient $\gamma$.}
		\label{tab:gamma_sensitivity}
        \setlength{\tabcolsep}{12pt}{
				\begin{tabular}{c|c|c|c|c}
					\hline
					\multirow{2}{*}{$\gamma$} & \multicolumn{2}{c|}{IEMOCAP} & \multicolumn{2}{c}{MELD} \\
					\cline{2-5}
					& w-Acc & w-F1 & w-Acc & w-F1 \\
					\hline
					0.0 & 73.47 & 72.88 & 65.88 & 66.08 \\
					0.2 & 74.45 & 74.06 & 66.35 & 66.20 \\
					0.4 & 74.63 & 74.25 & 66.27 & 66.39 \\
					0.6 & 74.79 & 74.44 & 66.49 & 66.60 \\
					0.8 & 74.64 & 74.35 & 66.42 & 66.51 \\
				\textbf{1.0} & \textbf{75.17} & \textbf{74.87} & \textbf{66.52} & \textbf{66.62} \\
					\hline
				\end{tabular}}
	\end{table}

	\textbf{Comparison with Denoising Baselines}
	To rule out the possibility that the gains mainly come from simple smoothing, we compare our method with moving average (MA), exponential moving average (EMA), and median filtering (Median). The results are shown in Table~\ref{tab:denoise_compare}. While these conventional methods can reduce noise to some extent, they consistently underperform our approach on both datasets. A key reason is that value-domain smoothing tends to attenuate emotion-relevant transient dynamics together with noise. In contrast, our method cancels steady relation redundancy in the relation domain and suppresses random jumps via gating, achieving a better trade-off between noise reduction and preservation of discriminative dynamics.
	
	\begin{table}[!t]
		\centering
		\caption{Comparison with conventional denoising strategies.}
		\label{tab:denoise_compare}
		\begin{threeparttable}
			\resizebox{\linewidth}{!}{%
            \setlength{\tabcolsep}{12pt}{
				\begin{tabular}{lcc|cc}
					\toprule
					\multirow{2}{*}{Method} & \multicolumn{2}{c|}{IEMOCAP} & \multicolumn{2}{c}{MELD} \\
					& w-Acc & w-F1 & w-Acc & w-F1 \\
					\midrule
					MA     & 73.44 & 73.15 & 64.28 & 64.25 \\
					EMA    & 74.20 & 73.86 & 65.87 & 65.78 \\
					Median & 73.73 & 73.54 & 66.14 & 65.92 \\
					\textbf{Ours}   & \textbf{75.17} & \textbf{74.87} & \textbf{66.52} & \textbf{66.62} \\
					\bottomrule
				\end{tabular}%
                }
			}
		\end{threeparttable}
	\end{table}

	\textbf{Comparison with Representative Attention Variants}
	To further clarify the differences between our method and existing difference-based attention mechanisms in terms of modeling formulation and underlying mechanism, we conduct controlled comparisons in which the overall architecture is kept unchanged and only the relation-domain differential operator is replaced with three representative variants, namely temporal difference attention, contrastive attention, and delta-attention. The results are reported in Table~\ref{tab:diffattn_compare}. Our model consistently achieves better performance across metrics, suggesting that the proposed relation-domain differencing and gating mechanism can suppress non-stationary noise while better preserving emotion-discriminative dynamics, thereby yielding a more favorable trade-off.
	
	\begin{table}[!t]
		\centering
		\caption{Comparison with representative difference-based attention variants under controlled settings.}
		\label{tab:diffattn_compare}
		\begin{threeparttable}
			\resizebox{\linewidth}{!}{%
            \setlength{\tabcolsep}{12pt}{
				\begin{tabular}{lcc|cc}
					\toprule
					\multirow{2}{*}{Method} & \multicolumn{2}{c|}{IEMOCAP} & \multicolumn{2}{c}{MELD} \\
					& w-Acc & w-F1 & w-Acc & w-F1 \\
					\midrule
					Temporal diff attn   & 72.18 & 72.12 & 65.55 & 65.76 \\
					Contrastive attn     & 74.42 & 73.48 & 65.69 & 65.77 \\
					Delta-attention      & 74.18 & 73.41 & 65.63 & 65.68 \\
					\textbf{Ours}        & \textbf{75.17} & \textbf{74.87} & \textbf{66.52} & \textbf{66.62} \\
					\bottomrule
				\end{tabular}%
                }
			}
		\end{threeparttable}
	\end{table}

	\textbf{Comparison of Fusion Weighting Strategies}
We compare the fixed-prior fusion strategy with two learnable fusion alternatives, LSF and ADF, under three text conditions: Normal, Mask 0.6, and Missing text. Table~\ref{tab:fusion_text_conditions} reports w-F1. The results show that the fixed-prior fusion performs best under all settings on MELD, and remains optimal under Normal and Mask 0.6 on IEMOCAP; however, under the Missing text setting on IEMOCAP, LSF and ADF achieve higher scores. This trend is consistent with our design motivation: during inference, the fixed fusion uses dataset-level weights to provide global calibration for the overall output magnitude of different modality branches, while sample-level complementary information is mainly modeled and captured by the feature-level cross-modal diffusion and gating modules.
\begin{table}[!t]
	\centering
	\caption{ w-F1 comparison of fixed-prior and learnable fusion strategies under different text conditions on MELD and IEMOCAP.}
	\label{tab:fusion_text_conditions}
	\begin{threeparttable}
		\setlength{\tabcolsep}{8pt}
		\begin{tabular}{llccc}
			\toprule
			Dataset & Text condition & Fixed & LSF & ADF \\
			\midrule
			\multirow{3}{*}{MELD}
			& $\mathrm{T0}$         & \textbf{66.62} & 65.19 & 66.03 \\
			& $\mathrm{T0.6}$       & \textbf{46.28} & 43.51 & 43.96 \\
			& $\mathrm{T\emptyset}$ & \textbf{34.73} & 33.55 & 33.26 \\
			\midrule
			\multirow{3}{*}{IEMOCAP}
			& $\mathrm{T0}$         & \textbf{74.87} & 73.24 & 72.79 \\
			& $\mathrm{T0.6}$       & \textbf{68.61} & 67.39 & 67.57 \\
			& $\mathrm{T\emptyset}$ & 27.22 & \textbf{35.68} & 32.71 \\
			\bottomrule
		\end{tabular}
		\begin{tablenotes}[para,flushleft]
			\footnotesize
			\textit{Note:} $\mathrm{T0}$ denotes clean text with no degradation;
			$\mathrm{T0.6}$ denotes text masking at evaluation with a masking ratio of 0.6;
			$\mathrm{T\emptyset}$ denotes the text-removed setting at evaluation, where textual inputs are entirely removed.
		\end{tablenotes}
	\end{threeparttable}
\end{table}

\textbf{Higher-order Reference Ablation}
	To further assess the practical sufficiency of the adjacent reference ($\Delta=1$) in differential attention, we extend the reference shift to $\Delta \in \{1,2,3\}$ while keeping all other training settings unchanged. As shown in Table \ref{tab:delta_ablation}, $\Delta=1$ achieves the best results on both IEMOCAP and MELD, whereas $\Delta=2$ and $\Delta=3$ do not bring consistent gains and instead show slight performance drops. This phenomenon suggests that more distant temporal references are more likely to cross true emotion transition points, thereby weakening useful short-term discriminative dynamics. Therefore, using two attention maps and constructing a first-order contrast with $\Delta=1$ is a more robust and efficient choice.

\begin{table}[H]
	\centering
	\caption{ Ablation on the temporal reference shift $\Delta$ in differential attention.}
	\label{tab:delta_ablation}
	\resizebox{\linewidth}{!}{%
		\begin{threeparttable}
			\setlength{\tabcolsep}{12pt}{
				\begin{tabular}{l|cc|cc}
					\hline
					\multirow{2}{*}{Reference shift $\Delta$} & \multicolumn{2}{c|}{IEMOCAP} & \multicolumn{2}{c}{MELD} \\
					\cline{2-5}
					& w-Acc (\%) & w-F1 (\%) & w-Acc (\%) & w-F1 (\%) \\
					\hline
					$\Delta=1$ (default) & \textbf{75.17} & \textbf{74.87} & \textbf{66.52} & \textbf{66.62} \\
					$\Delta=2$           & 74.30          & 74.14          & 66.36          & 66.50          \\
					$\Delta=3$           & 74.62          & 74.35          & 66.46          & 66.57          \\
					\hline
			\end{tabular}}
			\begin{tablenotes}[para,flushleft]
				\footnotesize
				$\Delta=1$ uses the adjacent (one-step) reference; other settings only change the reference shift.
			\end{tablenotes}
		\end{threeparttable}
	}
\end{table}

	\subsection{Case Study}
	
\begin{figure}[!t]
		\centering
		\includegraphics[width=\linewidth]{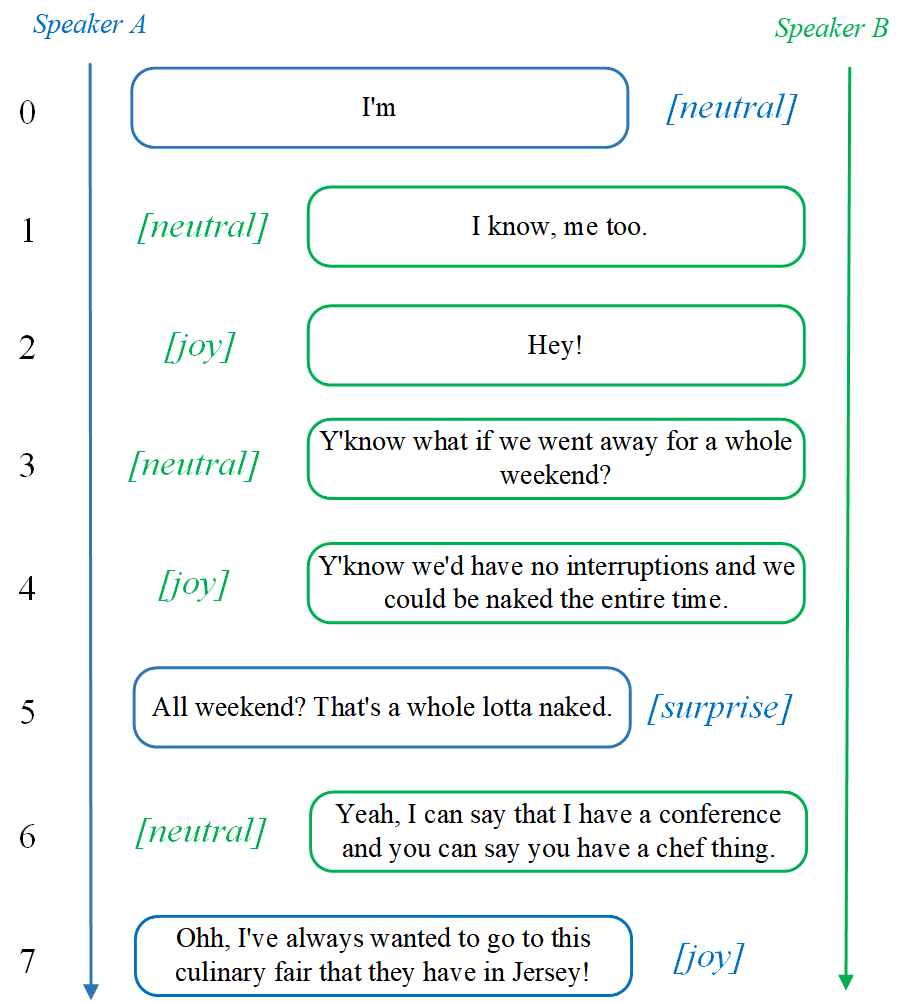}
		\caption{Illustration of the case dialogue window, including speaker turns, ground-truth emotion labels, and the target utterance position.}
		\label{fig:two1}
	\end{figure}

	As shown in Figure \ref{fig:two1}, we provide a qualitative visualization on an illustrative two-speaker window sampled from the MELD test set. Taking the final utterance in the window (local index $i=7$) as the prediction target, we visualize the behaviors of the differential denoising, relational graph interaction, and text-guided diffusion fusion modules under the same context.
	
\begin{figure}[!t]
		\centering
		\includegraphics[width=\linewidth]{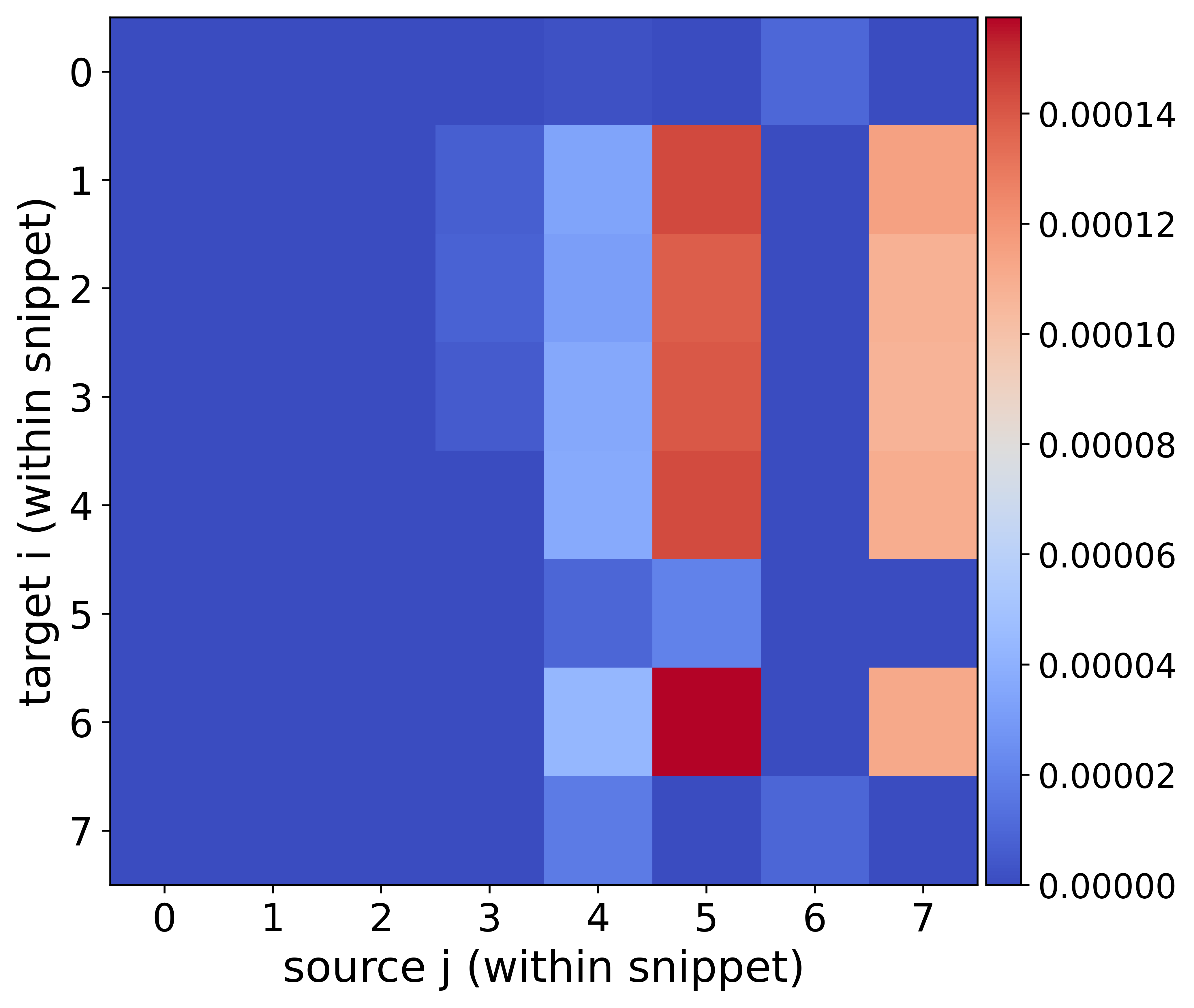}
		\caption{Heatmap of the acoustic relational-domain differential response ($\alpha - \lambda\alpha^{ref}$). Rows and columns correspond to the query position $i$ and the attended position $j$ within the local window, respectively; color intensity indicates the strength of relational variation relative to the temporal reference view.}
		\label{fig:audiodiff}
	\end{figure}

	\textbf{Differential Denoising} Figure \ref{fig:audiodiff} visualizes the acoustic relational-domain differential response produced by the differential denoising module. Most entries are close to zero, while a small set of locations exhibit pronounced responses, suggesting that relational differencing tends to suppress relatively stationary attention patterns and retain salient relational variations that are more likely correlated with emotion-discriminative dynamics.
	
\begin{figure}[!t]
		\centering
		\begin{subfigure}[b]{0.48\columnwidth}
			\centering
			\includegraphics[width=\textwidth]{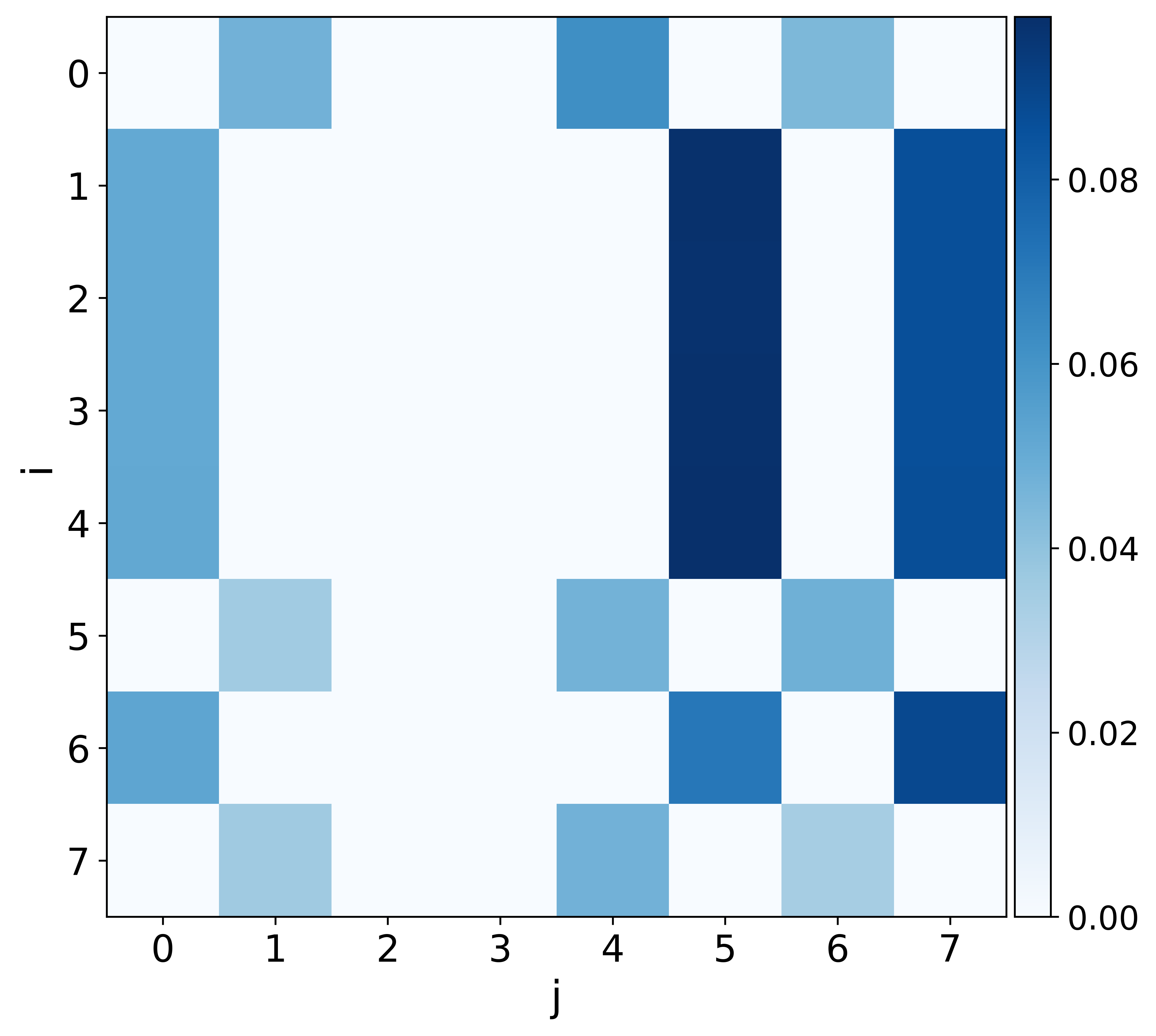}
			\caption{inter-speaker}
			\label{fig:heatmap1}
		\end{subfigure}
		\hfill
		\begin{subfigure}[b]{0.48\columnwidth}
			\centering
			\includegraphics[width=\textwidth]{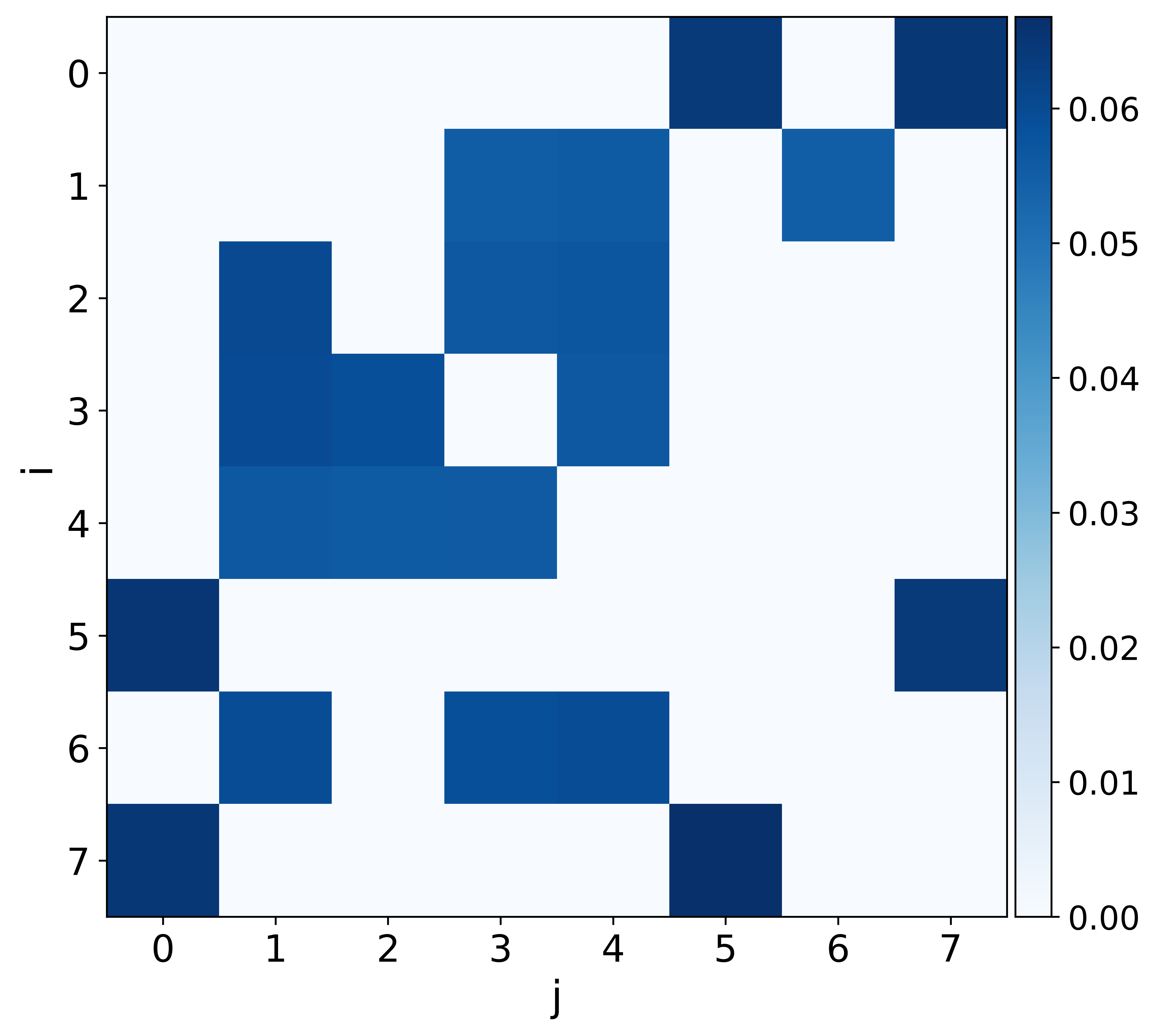}
			\caption{intra-speaker}
			\label{fig:heatmap2}
		\end{subfigure}
		\caption{Graph connectivity patterns on a two-speaker window from MELD. In each heatmap, row $i$ denotes the query utterance and column $j$ denotes the source utterance; darker colors indicate larger normalized weights.}
		\label{fig:overall}
	\end{figure}

	\textbf{Relational Graph Interaction} Figure  \ref{fig:overall} shows the edge-weight heatmaps of the inter-speaker and intra-speaker subgraphs (self-loops removed for clarity). The inter-speaker subgraph highlights influence pathways associated with the interlocutor’s utterances, whereas the intra-speaker subgraph captures continuity within the same speaker’s history, indicating a structured separation of inter- and intra-speaker dependencies.

\begin{figure}[!t]
		\centering
		\includegraphics[width=\linewidth]{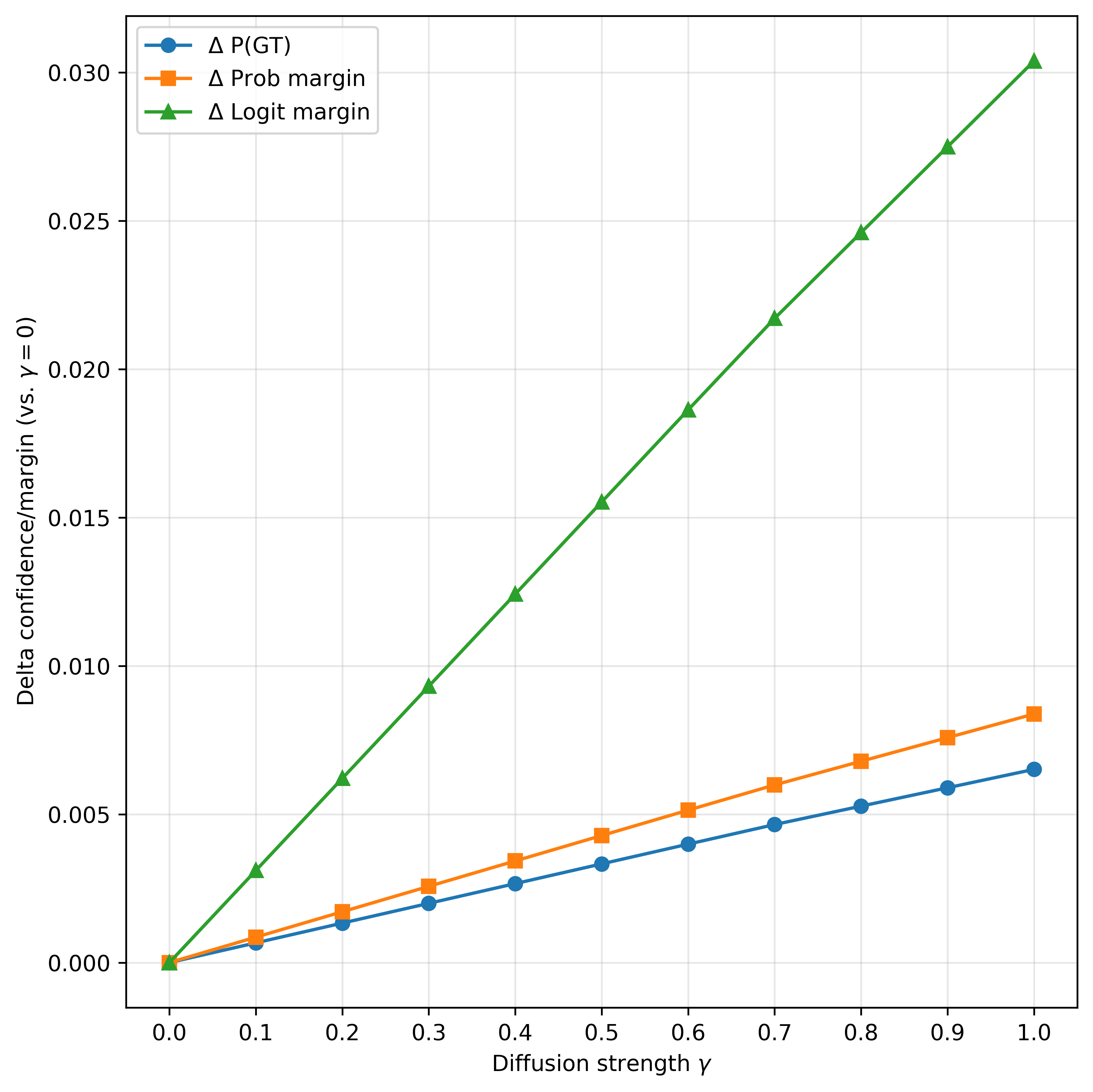}
		\caption{Incremental trajectories of target-utterance predictions under varying diffusion intensities $\gamma$ (relative to $\gamma=0$). $\Delta P(\mathrm{GT})$ denotes the increment of the ground-truth probability; $\Delta$probability margin denotes the increment of the probability difference between the ground-truth class and the strongest non-ground-truth class; $\Delta$logit margin denotes the increment of the corresponding logit difference.}
		\label{fig:melddifftrajconfdelta}
	\end{figure}

	\textbf{Text-guided Diffusion Attention Fusion} Figure \ref{fig:melddifftrajconfdelta} reports the incremental trends of prediction confidence and class margins for the target utterance relative to $\gamma=0$ as the diffusion intensity $\gamma$ varies. As $\gamma$ increases, $\Delta P(\mathrm{GT})$, $\Delta$probability margin, and $\Delta$logit margin exhibit an overall upward trend, indicating that text-guided diffusion fusion progressively injects acoustic and visual evidence into the text-dominant representation with controllable strength, thereby improving confidence and widening the separation from competing classes.
	
\subsection{Selective Denoising Analysis}

\begin{figure*}[!t]
	\centering
	\subfloat[Emotion-rich case]{
		\includegraphics[width=0.47\textwidth]{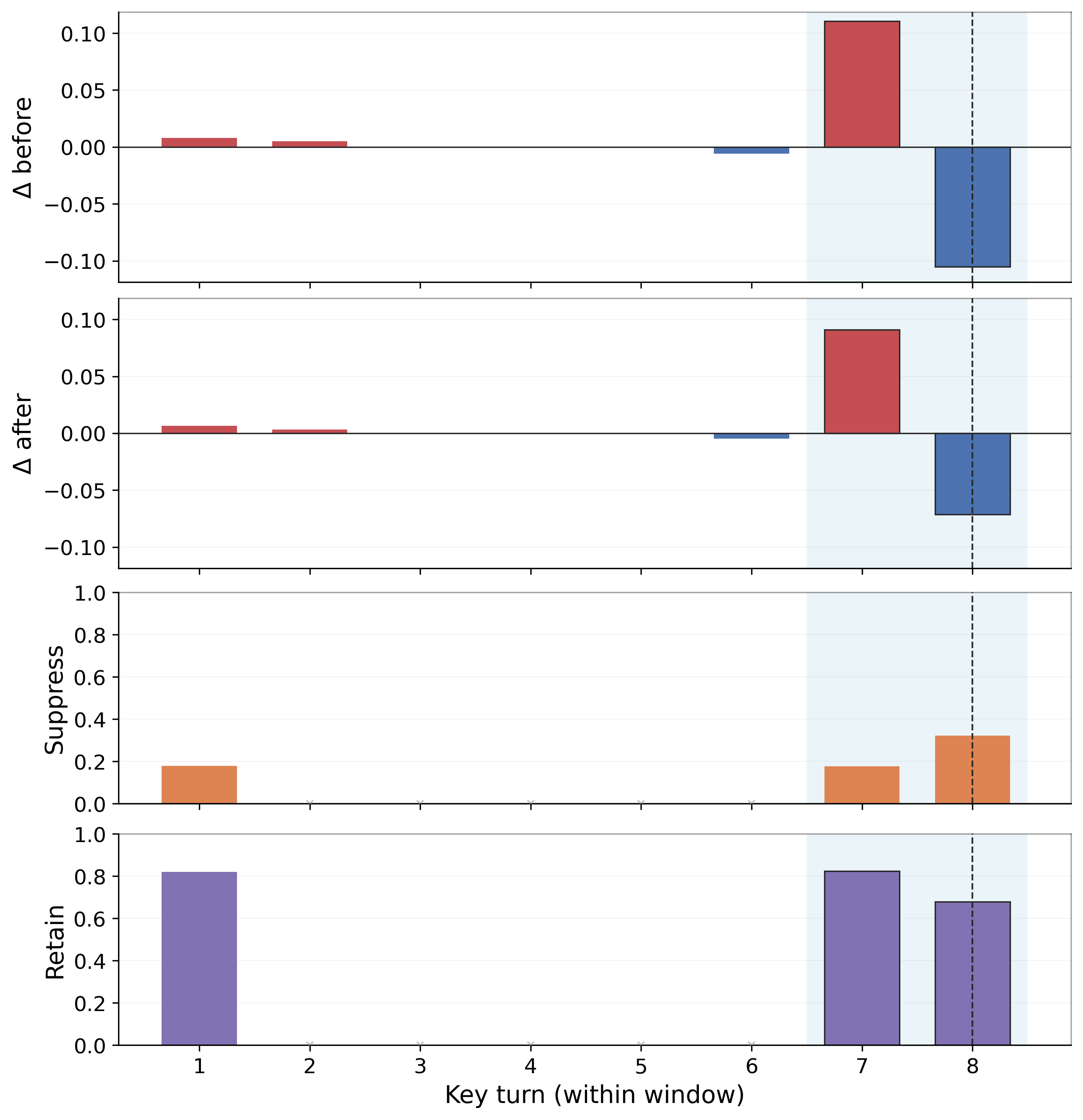}
	}
	\hfill
	\subfloat[Noisy-like case]{
		\includegraphics[width=0.47\textwidth]{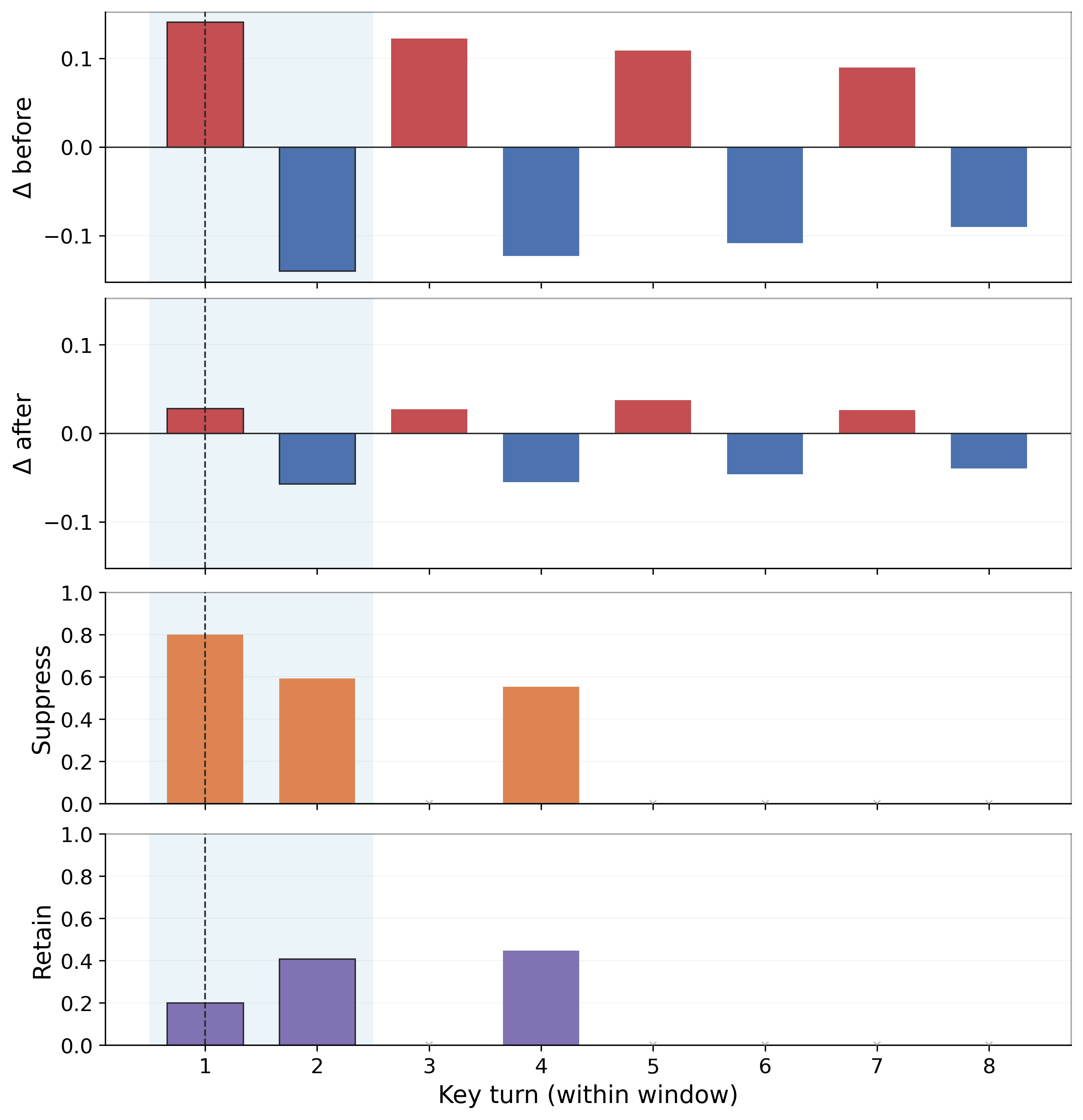}
	}
	\caption{Visualization of differential responses before and after gating at the target time step in the visual modality for an emotion-rich case (a) and a noisy-like case (b). In each panel, the four rows (top to bottom) show $\Delta_{\text{before}}$, $\Delta_{\text{after}}$, Suppress, and Retain. The dashed line marks the target key position, and the blue shaded region marks temporally neighboring key positions.}
	\label{fig:selective_denoising_visual_cases}
\end{figure*}

To more directly demonstrate the selective behavior of the differential denoising module, we visualize the differential responses at the target time step before and after gating in the visual modality. As shown in Figure \ref{fig:selective_denoising_visual_cases}, we present the pre-gating response $\Delta_{\text{before}}$, the post-gating response $\Delta_{\text{after}}$, the suppression map denoted as Suppress, and the retain-ratio map denoted as Retain. The two cases (emotion-rich and noisy-like) are determined automatically using a unified rule based on differential-response strength and local concentration statistics.

A comparison of the response row corresponding to the target time step shows that the gating operation does not apply a uniform scaling to all differential responses. Instead, it exhibits position-dependent filtering behavior across key positions. In emotion-rich segments, some salient responses remain with relatively high retain ratios after gating, whereas in noisy-like segments, multiple strong responses are clearly suppressed. Furthermore, under the same analysis setting, we compute retain-ratio statistics on 30 emotion-rich segments and 30 noisy-like segments in the visual modality. A two-sided Mann--Whitney U test shows a significant difference in retain ratio between the two groups, with $p=0.0119$ and Cliff's $\delta=0.378$. Specifically, the mean retain ratio of emotion-rich segments is 0.352 (standard deviation 0.107), which is higher than the mean retain ratio of noisy-like segments, 0.253 (standard deviation 0.085). These results indicate that the differential responses after gating exhibit differential retention patterns, which is consistent with the design goal of selective denoising.

	\section{Conclusion}
	In this study, we propose a multimodal emotion recognition framework that integrates a Differential Transformer, Relational Sub, and Cross-modal Diffusion Attention Fusion to address key challenges in conversational scenarios, including dynamic emotion tracking, noise interference in audio-visual modalities, and modality imbalance. The proposed framework employs a differential mechanism to effectively suppress noise in audio and visual features, leverages relational subgraphs to model both inter- and intra-speaker emotional dependencies, and performs fine-grained feature interactions through a text-guided cross-modal diffusion process. This approach fills the gap in existing research regarding modality denoising and text-centric interaction modeling, thereby significantly enhancing the robustness and expressive power of multimodal emotion recognition systems.
	
	Future work will explore approaches to model temporal dependencies in long conversations and further investigate the intrinsic correlations between multimodal signals and emotional states.

	\section*{CRediT authorship contribution statement}
	\textbf{Ying Liu}: Conceptualization, Methodology, Investigation, Data curation, Writing - Original Draft. 
	\textbf{Yuntao Shou}: Supervision, Investigation \& Review. 
	\textbf{Wei Ai}: Supervision, Investigation, Writing - Review \& Editing. 
	\textbf{Tao Meng}: Supervision, Investigation, Writing - Review \& Editing. 
	\textbf{Keqin Li}: Supervision, Investigation, Writing - Review \& Editing.

	\section*{Declaration of Competing Interest}
	The authors declare that they have no known competing financial interests or personal relationships that could have appeared to influence the work reported in this paper.
	
	\section*{Data availability}
	Data will be made available on request.
	
	\section*{Acknowledgements}
	\textcolor{black} {The authors deepest gratitude goes to the anonymous reviewers and AE for their careful work and thoughtful suggestions that have helped improve this paper substantially.} This work is supported by National Natural Science Foundation of China (Grant No. 69189338), Excellent Young Scholars of Hunan Province of China (Grant No. 22B0275), and program of Research on Local Community Structure Detection Algorithms in Complex Networks (Grant No. 2020YJ009).


	
	
	\bibliographystyle{unsrtnat}  
	\bibliography{ref}  


\end{document}